\documentclass{article} 
\usepackage{iclr2017_conference,times}
\usepackage{hyperref}
\usepackage{url}
\usepackage{amssymb}
\usepackage{amsmath}
\usepackage{tikz}
\usetikzlibrary{shapes.geometric,decorations.pathreplacing,arrows,fit,decorations.markings}
\usepackage{pgfplots}
\usepackage{graphicx}
\usepackage{multirow}
\usepackage{caption}
\usepackage{subcaption}
\usepackage{comment}
\usepackage{floatrow}
\usepackage{makecell}

\newfloatcommand{capbtabbox}{table}[][\FBwidth]
\iclrfinalcopy
\tikzstyle{curvebox} = [rectangle, minimum width=3cm, minimum height=1cm, text centered]
\tikzstyle{NN} = [rectangle,rounded corners,minimum width=4cm, minimum height=4cm, text centered, draw=black, fill=blue!30]
\tikzstyle{Cost} = [circle, inner sep = 7.5, draw=black, fill=red!30]
\tikzstyle{textbox} = [rectangle, minimum width=3cm, minimum height=1cm, text centered]
\tikzstyle{arrow} = [thick,->,>=latex, thick]
\tikzstyle{doublearrow} = [thick,<->,>=latex, thick, double,line width=1.75pt]
\newcommand\aaa{2cm}
\newcommand\CurveN{17cm}

\tikzstyle{Conv} = [rectangle, minimum width=2.5cm, minimum height=4cm, text centered,draw=black, fill=blue!30]
\tikzstyle{ReLU} = [rectangle, minimum width=2.5cm, minimum height=2.5cm, text centered,draw=black, fill=green!30]
\tikzstyle{CurveArray} = [rectangle, minimum width=0.1cm, minimum height=\CurveN, text centered,draw=black, fill=red!30]
\tikzstyle{CurveFatArray} = [rectangle, minimum width=1cm, minimum height=\CurveN, text centered,draw=black, fill=red!30]
\tikzstyle{Max} = [rectangle, minimum width=2.5cm, minimum height=2.5cm, text centered,draw=black, fill=yellow!30]
\tikzstyle{Linear} = [rectangle, minimum width=3cm, minimum height=3cm, text centered,draw=black,text width=1.95cm,fill=orange!30]

\title{\centering Learning Invariant Representations Of \\ Planar Curves}  

\author{Gautam Pai, Aaron Wetzler \& Ron Kimmel\\
Department of Computer Science\\
Technion-Israel Institute of Technology\\
\texttt{\{paigautam,twerd,ron\}@cs.technion.ac.il} \\
}

\begin{document}
\maketitle
\begin{abstract}
We propose a metric learning framework for the construction of invariant geometric
functions of planar curves for the Euclidean and Similarity group of transformations. 
We leverage on the representational power of convolutional neural
 networks to compute these geometric quantities. 
In comparison with axiomatic constructions, we show that the invariants
 approximated by the learning architectures have better numerical 
 qualities such as robustness to noise, resiliency to sampling, 
 as well as the ability to adapt to occlusion and partiality. 
Finally, we develop a novel \emph{multi-scale} representation in a similarity
 metric learning paradigm.  
\end{abstract}

\section{Introduction}
The discussion on \emph{invariance} is a strong component of the solutions to many
 classical problems in  numerical differential geometry. 
A typical example is that of planar shape analysis where one desires to have a local function of the contour 
 which is invariant to rotations, translations and reflections like the Euclidean curvature. 
This representation can be used to obtain correspondence between the shapes and also to compare and classify them. 
However, the numerical construction of such functions from discrete sampled data is non-trivial and requires robust 
 numerical techniques for their stable and efficient computation.

Convolutional neural networks have been very successful in recent years in solving
 problems in image processing, recognition and classification. 
Efficient architectures have been studied and developed to extract semantic features from images invariant to a certain
 class or category of transformations. 
Coupled with efficient optimization routines and more importantly, a large amount of data, 
 a convolutional neural network  can be trained to construct invariant representations and 
 semantically significant features of images as well as other types of data such as speech and language. 
It is widely acknowledged that such networks have superior representational power compared to more principled
 methods with more handcrafted features such as wavelets, Fourier methods, kernels etc. which are not optimal for 
 more semantic data processing tasks. 

\begin{figure}[t]
\vspace{-1cm}
\includegraphics[width=12cm, height = 5cm]{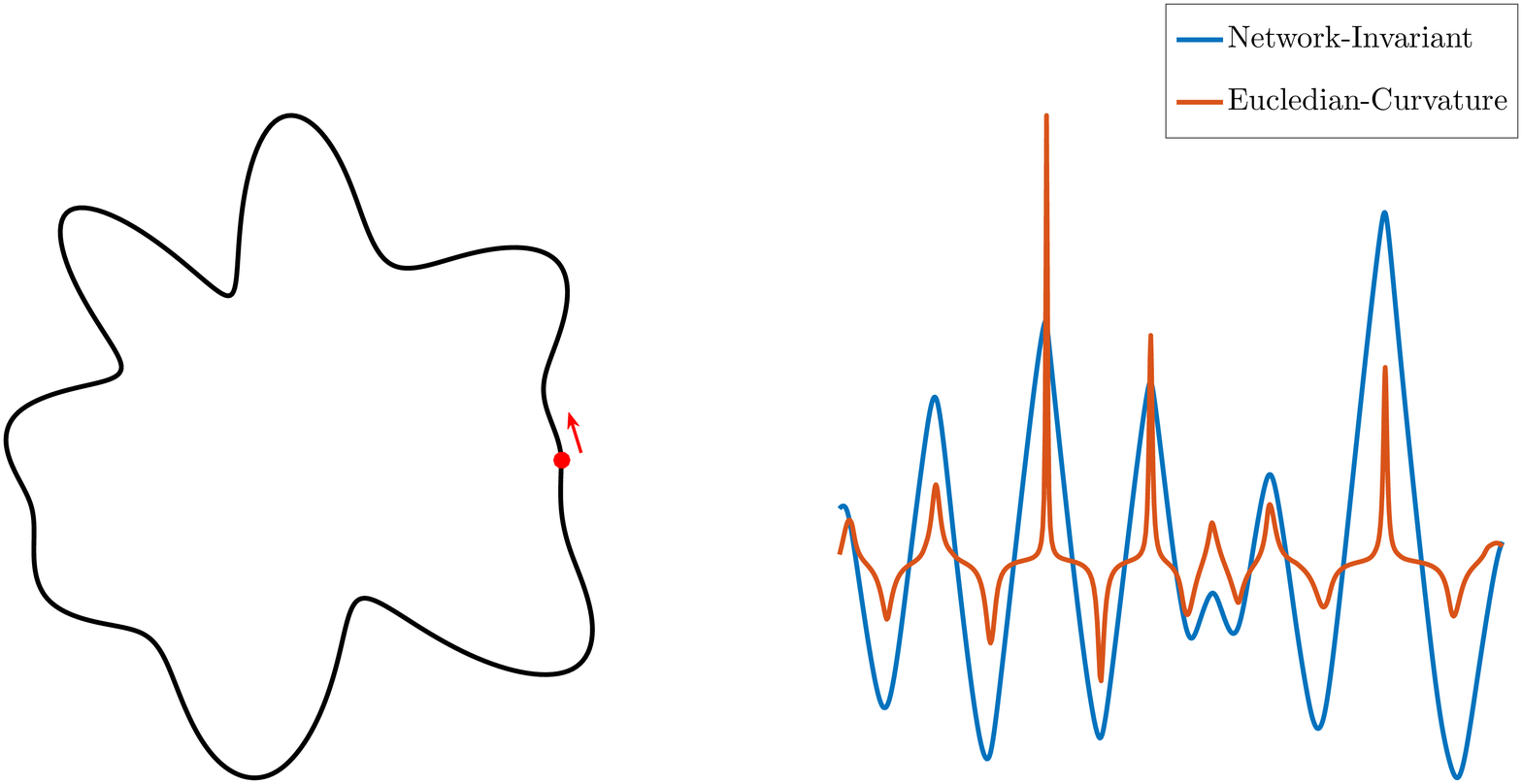}
\caption{Comparing the axiomatic and learned invariants of a curve.}
\label{Figure1}
\end{figure}

In this paper we connect two seemingly different fields: convolutional neural network based metric
 learning methods and numerical differential geometry. 
The results we present are the outcome of investigating the question: 
 \emph{"Can metric learning methods be used to construct invariant geometric quantities?"}    
By training with a Siamese configuration involving only positive and negative examples of 
 Euclidean transformations,
 we show that the network is able to train for an invariant geometric function of the curve which can be contrasted
 with a theoretical quantity: Euclidean curvature. 
An example of each can be seen Figure \ref{Figure1}. 
We compare the learned invariant functions with axiomatic counterparts and provide a discussion 
 on their relationship. 
Analogous to principled constructions like curvature-scale space methods and integral invariants, 
 we develop a multi-scale representation using a data-dependent learning based approach. 
We show that network models are able 
 to construct geometric invariants that are numerically more stable and robust than these 
 more principled approaches. 
We contrast the computational work-flow of a typical numerical geometry pipeline with that of 
 the convolutional neural network model and develop a relationship among them highlighting 
 important geometric ideas.

In Section \ref{Background} we begin by giving a brief summary of the theory and history of 
 invariant curve representations. 
In Section \ref{TrainingNet} we explain our main contribution of casting the problem into the form which enables
training a convolutional neural network for generating invariant signatures to the Euclidean and Similarity group transformations. 
Section \ref{MultiScale} provides a discussion on developing a multi-scale representation followed
 by the experiments and discussion in Section \ref{Experiments}.  
 
\section{Background}
\label{Background}
An invariant representation of a curve is the set of signature functions assigned to every point 
 of the curve which does not change despite the action of a certain type of transformation. 
A powerful theorem from E. Cartan (\cite{cartan1983geometry}) and Sophus Lie (\cite{ackerman1976sophus}) 
 characterizes the space of these invariant signatures. 
It begins with the concept of arc-length which is a generalized notion of the length along a curve. 
Given a type of transformation, one can construct an intrinsic arc-length that is
 independent of the parameterization of the curve, and compute the curvature with respect
 to this arc-length. 
The fundamental invariants of the curve, known as differential invariants 
 (\cite{bruckstein1995differential}, \cite{calabi1998differential}) are the set of functions 
  comprising of the curvature and its successive derivatives with respect to the invariant arc-length. 
These differential invariants are unique in a sense that two curves are related by the group
 transformation if and only if their differential invariant signatures are identical. 
Moreover, every invariant of the curve is a function of these fundamental differential invariants.   
Consider $C(p)=\begin{bmatrix} x(p)\\y(p)\end{bmatrix}$: a planar curve with coordinates 
 $x$ and $y$ parameterized by some parameter $p$. 
The Euclidean arc-length, is given by
\begin{equation}
s(p) = \int_0^p|C_p|\;dp = \int_0^p \sqrt{x_p^2 + y_p^2} \; dp,
\label{euc_arc}
\end{equation}
where $x_p = \frac{dx}{dp}$, and $y_p = \frac{dy}{dp}$ and the principal invariant signature, 
 that is the Euclidean curvature is given by 
\begin{equation}
\kappa(p) = \frac{\det(C_p, C_{pp})}{|C_p|^3} = \frac{x_p y_{pp} - y_px_{pp}}{(x_p^2+y_p^2)^\frac{3}{2}}.
\label{euc_curvature}
\end{equation}
Thus, we have the Euclidean differential invariant signatures given by the set 
 $\{\kappa,\;\kappa_s,\; \kappa_{ss}\;...\}$  for every point on the curve. 
Cartan's theorem provides an axiomatic construction of invariant signatures and the uniqueness 
 property of the theorem guarantees their theoretical validity. 
Their importance is highlighted from the fact that \emph{any} invariant is a function 
 of the fundamental differential invariants.

The difficulty with differential invariants is their stable numerical computation.  
Equations \ref{euc_arc} and \ref{euc_curvature}, involve non-linear functions of derivatives 
 of the curve and this poses serious numerical issues for their practical implementation where 
  noise and poor sampling techniques are involved. 
Apart from methods like \cite{pajdla1995matching} and \cite{weiss1993noise}, numerical 
 considerations motivated the development of multi-scale representations. 
These methods used alternative constructions of invariant signatures which were robust to noise. 
More importantly, they allowed a hierarchical representation, in which the strongest and the most
 global components of variation in the contour of the curve are encoded in signatures of higher scale, 
  and as we go lower, the more localized and rapid changes get injected into the representation. 
Two principal methods in this category are scale-space methods and integral invariants. 
In scale-space methods (\cite{mokhtarian1992theory}; \cite{sapiro1995area};
 \cite{bruckstein1996recognizing}), the curve is subjected to an invariant evolution process where
 it can be evolved to different levels of abstraction. 
See Figure \ref{curvature_flow}.
The curvature function at each evolved time $t$ is then recorded as an invariant. 
For example, $\{ \kappa(s,t), \kappa_s(s,t), \kappa_{ss} (s,t) ... \}$ would be the 
 Euclidean-invariant representations at scale $t$. 

Integral invariants (\cite{manay2004integral}; \cite{fidler2008identifiability}; \cite{pottmann2009integral};
 \cite{hong2015shape}) are invariant signatures which compute integral measures along the curve. 
For example, for each point on the contour, the integral area invariant computes the area of the 
 region obtained from the intersection of a ball of radius $r$ placed at that point and the interior 
  of the contour. 
The integral nature of the computation gives the signature robustness to noise and by adjusting 
 different radii of the ball $r$ one can associate a scale-space of responses for this invariant.
 \cite{fidler2008identifiability} and \cite{pottmann2009integral} provide a detailed treatise on 
  different types of integral invariants and their properties.  

It is easy to observe that differential and integral invariants can be thought of as being obtained
 from non-linear operations of convolution filters. 
The construction of differential invariants employ filters for which the action is equivalent to 
 numerical differentiation (high pass filtering) whereas integral invariants use filters which act 
  like numerical integrators (low pass filtering) for stabilizing the invariant. 
This provides a motivation to adopt a learning based approach and we demonstrate that the process 
 of estimating these filters and functions can be outsourced to a learning framework. 
We use the Siamese configuration for implementing this idea. 
Such configurations have been used in signature verification (\cite{bromley1993signature}), 
 face-verification and recognition(\cite{sun2014deep}; \cite{taigman2014deepface};
  \cite{hu2014discriminative}), metric learning (\cite{chopra2005learning}), image descriptors
   (\cite{carlevaris2014learning}), dimensionality reduction (\cite{hadsell2006dimensionality}) 
    and also for generating 3D shape descriptors for correspondence and retrieval 
     (\cite{masci2015geodesic}; \cite{xie2015deepshape}). 
In these papers, the goal was to learn the descriptor and hence the similarity metric from data using
 notions of only positive and negative examples. 
We use the same framework for estimation of geometric invariants. 
However, in contrast to these methods, we contribute an analysis of the output 
 descriptor and provide a geometric context to the learning process. 
The contrastive loss function driving the training ensures that the network chooses filters 
 which push and pull different features of the curve into the invariant by balancing a mix 
  of robustness and fidelity. 
\begin{figure}[t]
\vspace{-0.5cm}
\begin{tikzpicture}[thick, every node/.style={scale=0.25,font=\Huge}]
\node(textLeft)[textbox]{Curve1: $C_1$};
\node(textRight)[textbox,right of = textLeft, xshift = 13cm]{Curve2: $C_2$};
\node(Label)[textbox, left of = textLeft, xshift = -7cm]{Label: $\lambda \in \{0,1\}$};

\node (curveLeft) [curvebox, below of = textLeft, yshift = -1.5cm] {\includegraphics[scale = 0.16]{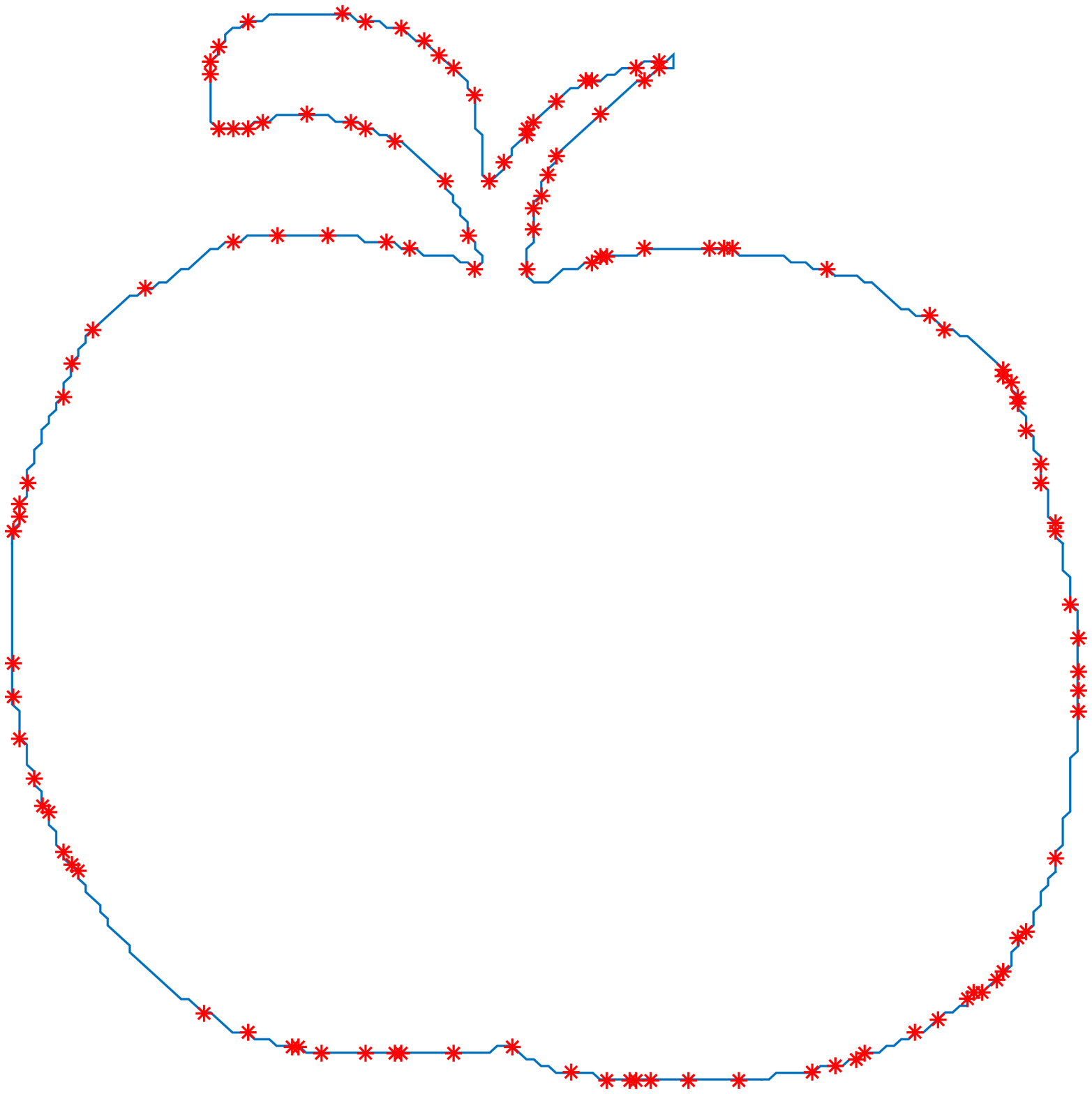}};
\node (curveRight)[curvebox, below of = textRight, yshift = -1.5cm] {\includegraphics[scale = 0.16]{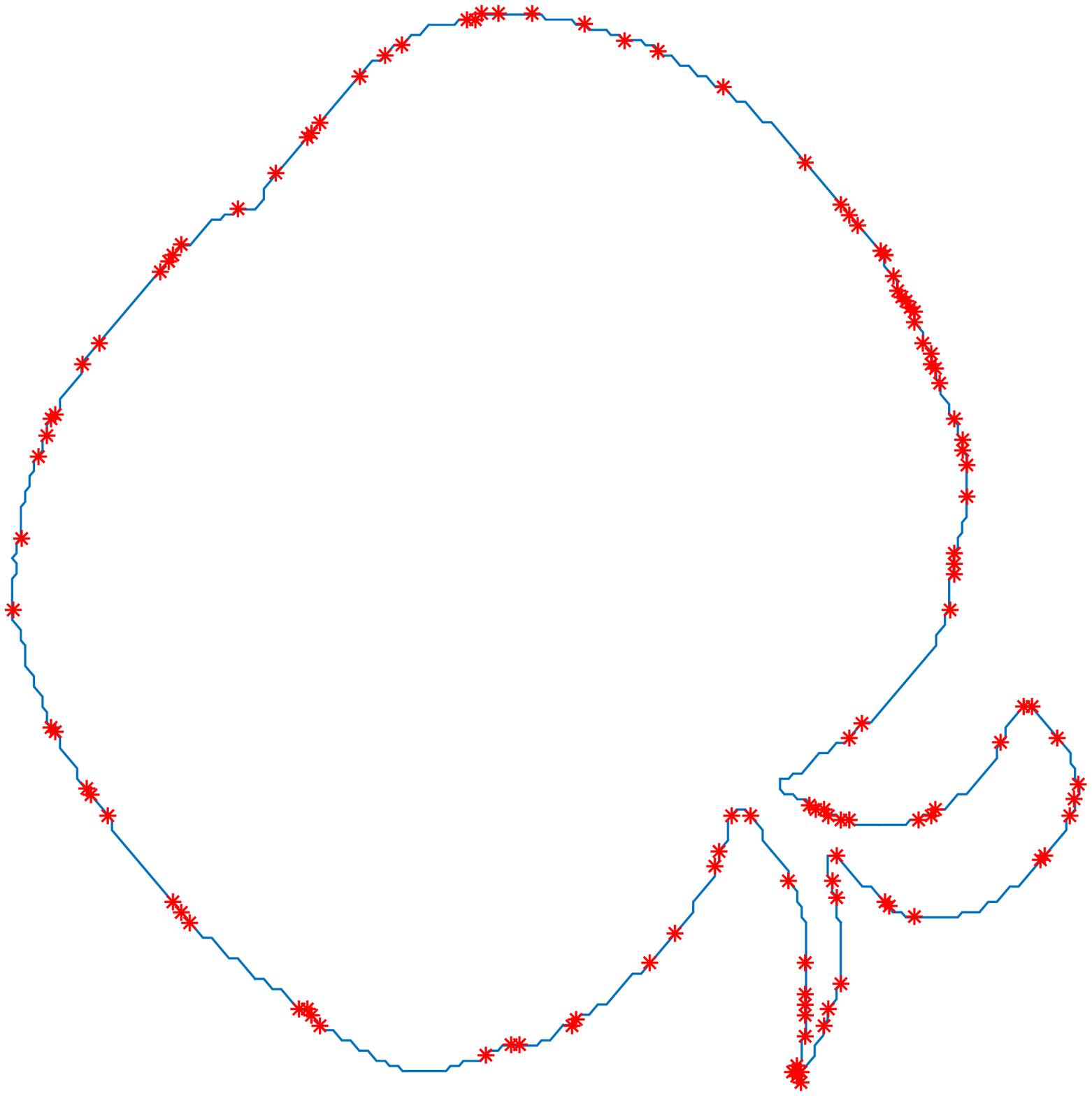}};

\node (NN_Left) [NN, below of  = curveLeft, yshift = -3.5cm] {Network1: $\Theta$};
\node (NN_Right) [NN, below of = curveRight, yshift = -3.5cm] {Network2: $\Theta$};

\node(textSigLeft)[textbox, below of = NN_Left, yshift = -3.5cm]{Output1: $\mathbf{\mathcal{S}_\Theta}(C_{1})$};
\node(textSigRight)[textbox,below of = NN_Right, yshift = -3.5cm]{Output2: $\mathbf{\mathcal{S}_\Theta}(C_{2})$};

\node (SigLeft) [curvebox, below of = textSigLeft, yshift = -1cm] {\includegraphics[width=15cm,height=2cm]{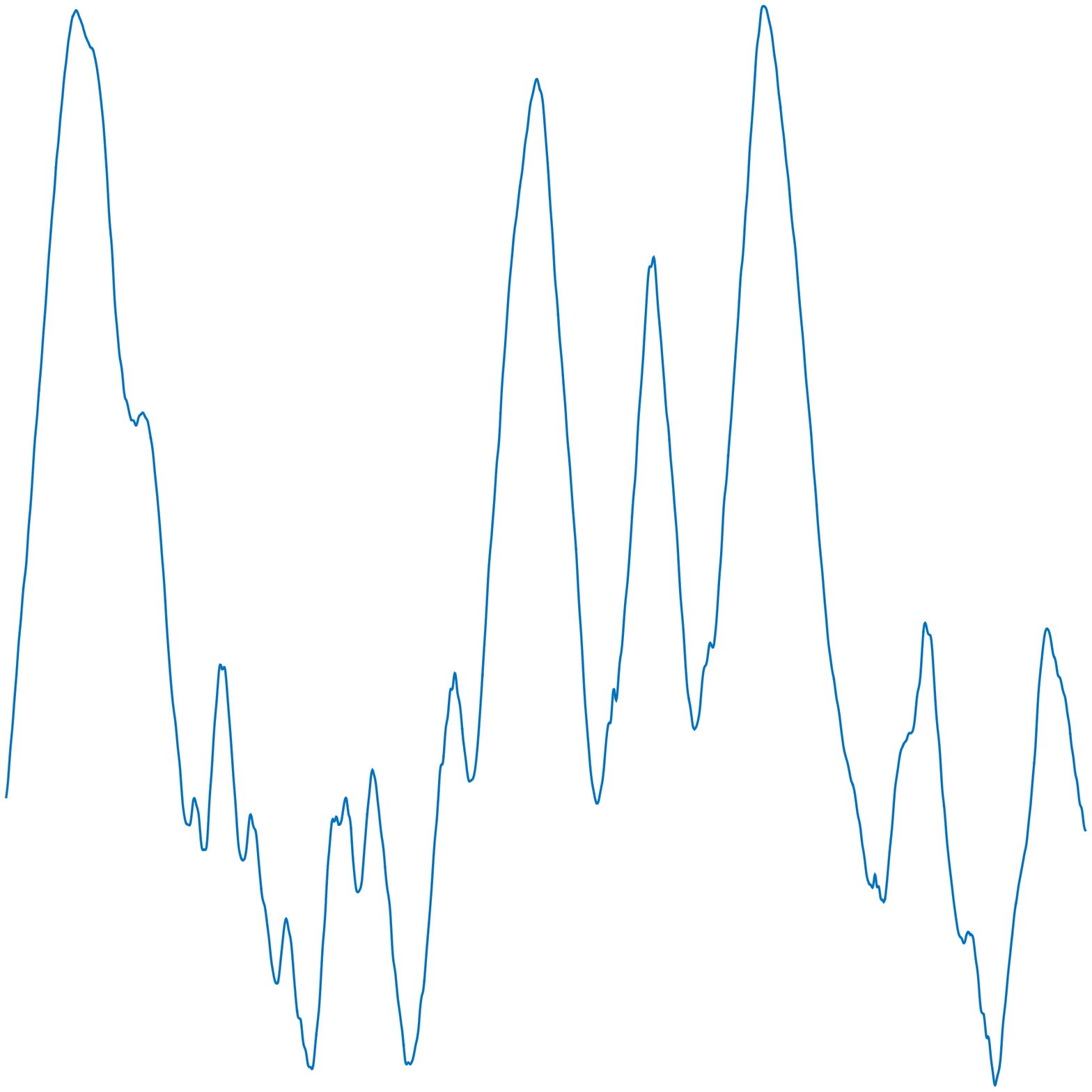}};
\node (SigRight)[curvebox, below of = textSigRight, yshift = -1cm] {\includegraphics[width=15cm,height=2cm]{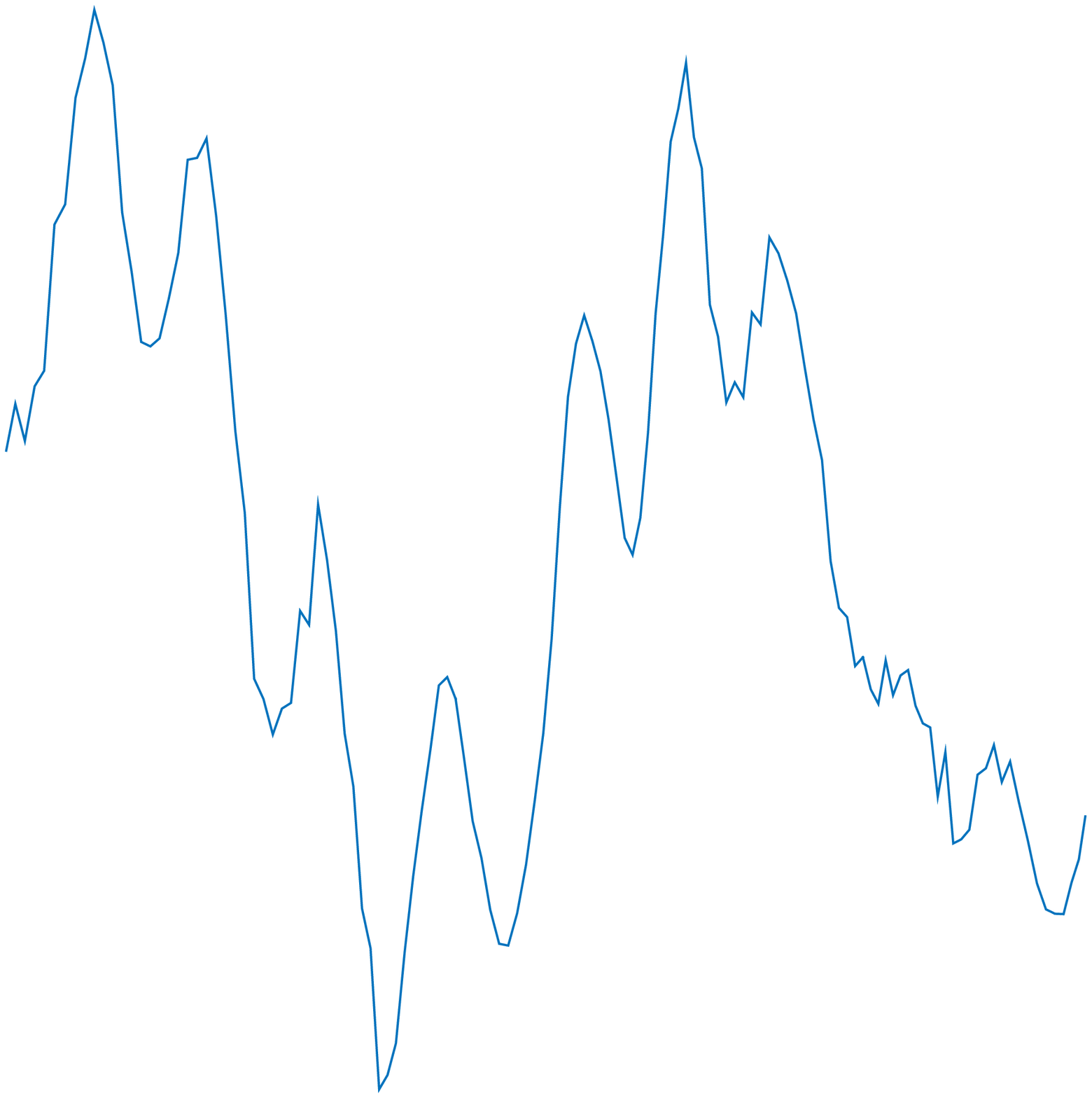}};

\node (Cost) [Cost, below of = SigLeft,yshift  = -5cm, xshift = 7cm]{Cost: $\mathcal{L}(\Theta)$};

\node(MathCost)[textbox, below of = Cost,yshift  = -4cm]{$\mathcal{L}(\Theta) = \;\lambda \;\; ||\; \mathcal{S}_\Theta(C_{1}) - \mathcal{S}_\Theta(C_{2})\;||  \;  \;+\;(1 - \lambda) \;\; \max (\;0,\;\mu \; - \; ||\; \mathcal{S}_\Theta(C_{1}) - \mathcal{S}_\Theta(C_{2})\;|| \;)$};

\draw [arrow] (NN_Left) -- (textSigLeft); 
\draw [arrow] (NN_Right) -- (textSigRight);

\draw [arrow] (curveLeft)  -- (NN_Left) ; 
\draw [arrow] (curveRight) -- (NN_Right);

\draw [arrow] (SigLeft)  -- (Cost); 
\draw [arrow] (SigRight) -- (Cost);

\draw [doublearrow] (NN_Left) -- node [anchor=south,above=0.5cm]{Shared Weights}(NN_Right);
\draw[arrow] (Label) |- (Cost);
\end{tikzpicture}
\caption{Siamese Configuration}
\label{Siamese}
\end{figure}
\section{Training For Invariance}
\label{TrainingNet}
A planar curve can be represented either explicitly by sampling points on the curve or  
 using an implicit representation such as level sets (\cite{kimmel2012numerical}). 
We work with an explicit representation of simple curves (open or closed) with random variable 
 sampling of the points along the curve. 
Thus, every curve is a $N\;\times\;2$ array denoting the $X$ and $Y$ coordinates of the $N$ points. 
We build a convolutional neural network which inputs a curve and outputs a representation or 
 signature for every point on the curve. 
We can interpret this architecture as an algorithmic scheme of representing a function over the curve.
However feeding in a single curve is insufficient and instead we run this convolutional architecture in 
a Siamese configuration (Figure \ref{Siamese}) that accepts a curve and a transformed version (positive) of the curve 
or an unrelated curve (negative). By using two identical copies
 of the same network sharing weights to process these two curves we are able to extract geometric invariance by using a loss function 
 to require that the two arms  of the Siamese configuration must produce values that are minimally different for curves
 which are related by Euclidean transformations representing positive examples and maximally different for carefully constructed negative examples. 
To fully enable training of our network we build a large dataset comprising of positive and negative examples of the 
 relevant transformations from a database of curves. 
We choose to minimize the contrastive loss between the two outputs of the Siamese network as this directs the network architecture
 to model a function over the curve which is invariant to the transformation.
\subsection{Loss Function}
We employ the contrastive loss function (\cite{chopra2005learning}; \cite{lecun2006tutorial}) 
 for training our network. 
The Siamese configuration comprises of two identical networks of Figure \ref{Architecture} 
 computing signatures for two separate inputs of data. 
Associated to each input pair is a label which indicates whether or not that pair is a 
 positive $(\lambda =1)$  or a negative $(\lambda = 0)$ example (Figure \ref{Siamese}). 
Let $C_{1i}$ and $C_{2i}$ be the curves imputed to first and second arm of the configuration
 for the $i^{th}$ example of the data with label $\lambda_i$. 
Let $\mathcal{S}_\Theta(C)$ denote the output of the network for a given set of weights $\Theta$ 
 for input curve $C$.
The contrastive loss function is given by:
\begin{equation}
\mathcal{C}(\Theta) = \frac{1}{N} \big{\{}\sum_{i=1}^{i=N} \;\lambda_i \;\; ||\; \mathcal{S}_\Theta(C_{1i}) - \mathcal{S}_\Theta(C_{2i})\;||  \;  \;+\;(1 - \lambda_i) \;\; \max (\;0,\;\mu \; - \; ||\; \mathcal{S}_\Theta(C_{1i}) - \mathcal{S}_\Theta(C_{2i})\;|| \;) \big{\}},
\label{siamese_loss}
\end{equation}
 where $\mu$ is a cross validated hyper-parameter known as {\it margin} which defines the metric threshold 
 beyond which negative examples are penalized.  
\subsection{Architecture}
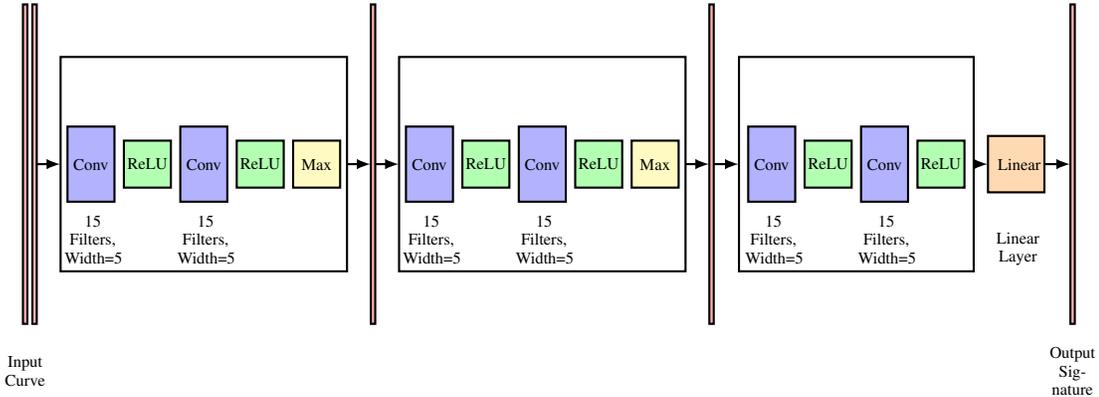
\begin{figure}[t]
\vspace{-0.5cm}
\begin{tikzpicture}[thick, every node/.style={scale=0.25,font=\Huge}]
\node(CurveX)[CurveArray]{};
\node(CurveY)[CurveArray, right of = CurveX, xshift=-0.5cm]{};
\node(textCurve)[textbox,below of = CurveX,yshift = -10cm,text width=3cm]{Input Curve};

\node(ConvL1_1)[Conv,right of = CurveY, xshift = 2cm]{Conv};
\node(textConv)[textbox, below of = ConvL1_1, text width=3cm, yshift= -3cm, xshift = 0.1cm]{15 Filters, Width=5};
\node(ReLUL1_1)[ReLU,right of = ConvL1_1,xshift = \aaa]{ReLU};
\node(ConvL1_2)[Conv,right of = ReLUL1_1,xshift = \aaa]{Conv};
\node(textConv)[textbox, below of = ConvL1_2, text width=3cm, yshift= -3cm, xshift = 0.1cm]{15 Filters, Width=5};
\node(ReLUL1_2)[ReLU,right of = ConvL1_2,xshift = \aaa]{ReLU};

\node(Max_L1)[Max,right of = ReLUL1_2,xshift = \aaa]{Max};
\node(Out_L1)[CurveArray, right of = Max_L1,xshift = 2cm]{};
\node [fit = (ConvL1_1)(ReLUL1_1)(ConvL1_2)(ReLUL1_2)(Max_L1),inner xsep = 5.8cm,inner ysep = 5.2cm,draw] (box1){};

\node(ConvL2_1)[Conv,right of = Out_L1,xshift = 2cm]{Conv};
\node(textConvL21)[textbox, below of = ConvL2_1, text width=3cm, yshift= -3cm, xshift = 0.1cm]{15 Filters, Width=5};
\node(ReLUL2_1)[ReLU,right of = ConvL2_1,xshift = \aaa]{ReLU};
\node(ConvL2_2)[Conv,right of = ReLUL2_1,xshift = \aaa]{Conv};
\node(textConvL22)[textbox, below of = ConvL2_2, text width=3cm, yshift= -3cm, xshift = 0.1cm]{15 Filters, Width=5};
\node(ReLUL2_2)[ReLU,right of = ConvL2_2,xshift = \aaa]{ReLU};
\node(Max_L2)[Max,right of = ReLUL2_2,xshift = \aaa]{Max};
\node(Out_L2)[CurveArray, right of = Max_L2,xshift = 2cm]{};
\node [fit = (ConvL2_1)(ReLUL2_1)(ConvL2_2)(ReLUL2_2)(Max_L2),inner xsep = 5.8cm,inner ysep = 5.2cm,draw] (box2){};

\node(ConvL3_1)[Conv,right of = Out_L2,xshift = 2.2cm]{Conv};
\node(textConvL31)[textbox, below of = ConvL3_1, text width=3cm, yshift= -3cm, xshift = 0.1cm]{15 Filters, Width=5};
\node(ReLUL3_1)[ReLU,right of = ConvL3_1,xshift = \aaa]{ReLU};
\node(ConvL3_2)[Conv,right of = ReLUL3_1,xshift = \aaa]{Conv};
\node(textConvL32)[textbox, below of = ConvL3_2, text width=3cm, yshift= -3cm, xshift = 0.1cm]{15 Filters, Width=5};
\node(ReLUL3_2)[ReLU,right of = ConvL3_2,xshift = \aaa]{ReLU};
\node [fit = (ConvL3_1)(ReLUL3_1)(ConvL3_2)(ReLUL3_2),inner xsep = 4.8cm,inner ysep = 5.2cm,draw] (box3){};
\node(Linear_L3)[Linear,right of = ReLUL3_2,xshift = 3cm]{Linear};
\node(Linear)[textbox, below of = Linear_L3, text width=3cm, yshift= -3.5cm, xshift = 0.1cm]{Linear Layer};
\node(Out_L3)[CurveArray, right of = Linear_L3,xshift = 2cm]{};
\node(textOut3)[textbox,below of = Out_L3,yshift = -10cm,text width=3cm]{Output Signature};

\draw [arrow] (CurveY) -- (box1); 
\draw [arrow] (box1) -- (Out_L1);
\draw [arrow] (Out_L1) -- (box2);
\draw [arrow] (box2) -- (Out_L2);
\draw [arrow] (Out_L2) -- (box3);
\draw [arrow] (box3) -- (Linear_L3);
\draw [arrow] (Linear_L3) -- (Out_L3);
\end{tikzpicture}
\caption{Network Architecture}
\label{Architecture}
\end{figure}
The network inputs a $N \times 2$ array representing the coordinates of $N$ points along the curve. 
The sequential nature of the curves and the mostly $1D$-convolution operations can also be looked at 
from the point of view of temporal signals using recurrent neural network architectures. Here however we
choose instead to use a multistage CNN pipeline.
The network, given by one arm of the Siamese configuration, comprises of three stages that use layer units
which are typically considered the basic building blocks of modern CNN architectures.
Each stage contains two sequential batches of convolutions appended with rectified linear
 units (ReLU) and ending with a max unit. The convolutional unit comprises of convolutions with $15$ filters of width $5$ as depicted in
 Figure \ref{Architecture}.
The max unit computes the maximum of $15$ responses per point to yield an intermediate output after
 each stage.  
The final stage is followed by a linear layer which linearly combines the responses to yield the final output. 
Since, every iteration of convolution results in a reduction of the sequence length, sufficient
 padding is provided on both ends of the curve. 
This ensures that the value of the signature at a point is the result of the response of the 
 computation resulting from the filter centered around that point. 
 
\begin{figure}[t]
\hspace{-0.4\textwidth}
\begin{subfigure}[h]{0.45\textwidth}
\centering
\includegraphics[scale=0.35]{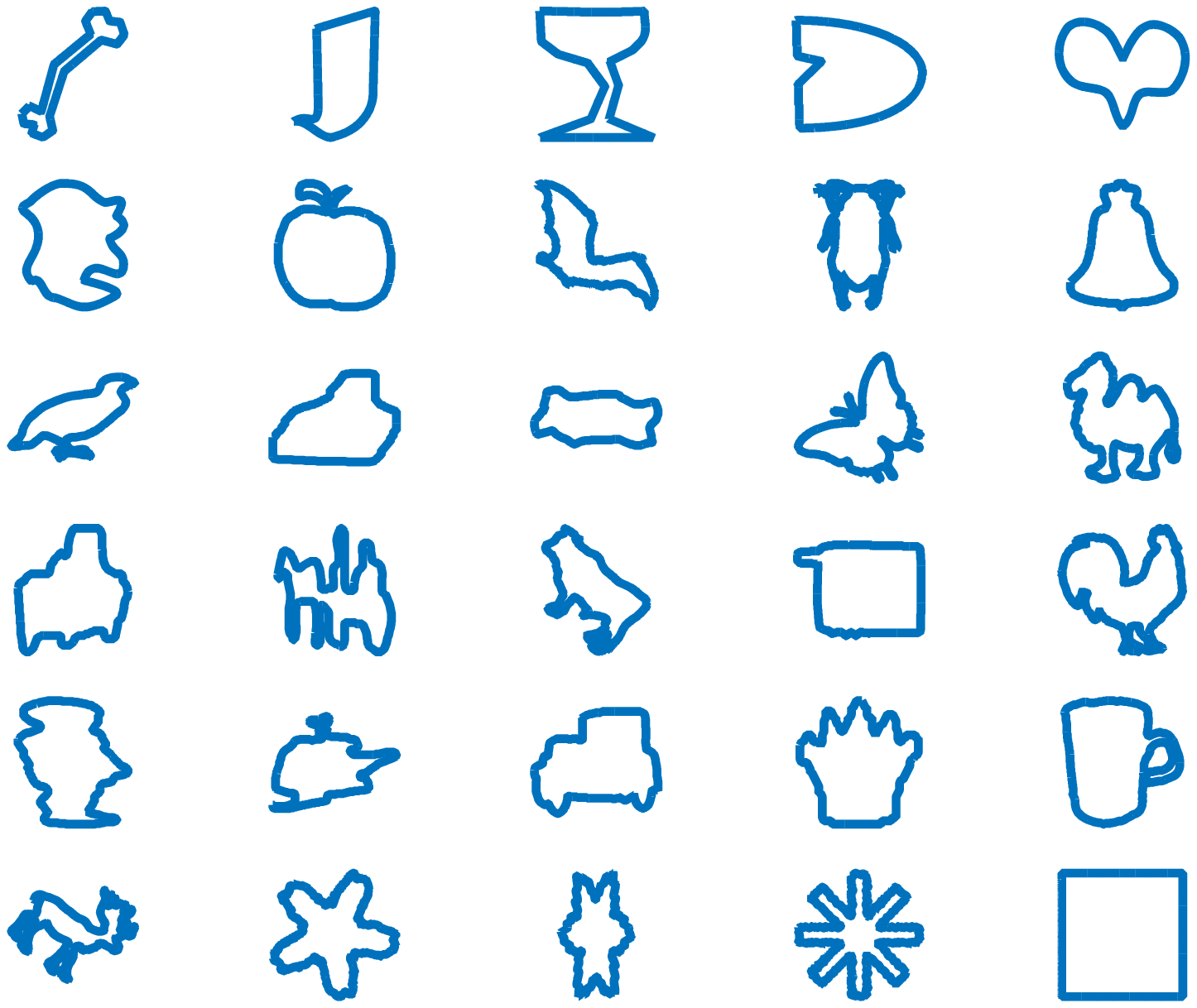}
\end{subfigure}
\begin{subfigure}[h]{0.3\textwidth}
\includegraphics[scale=0.21]{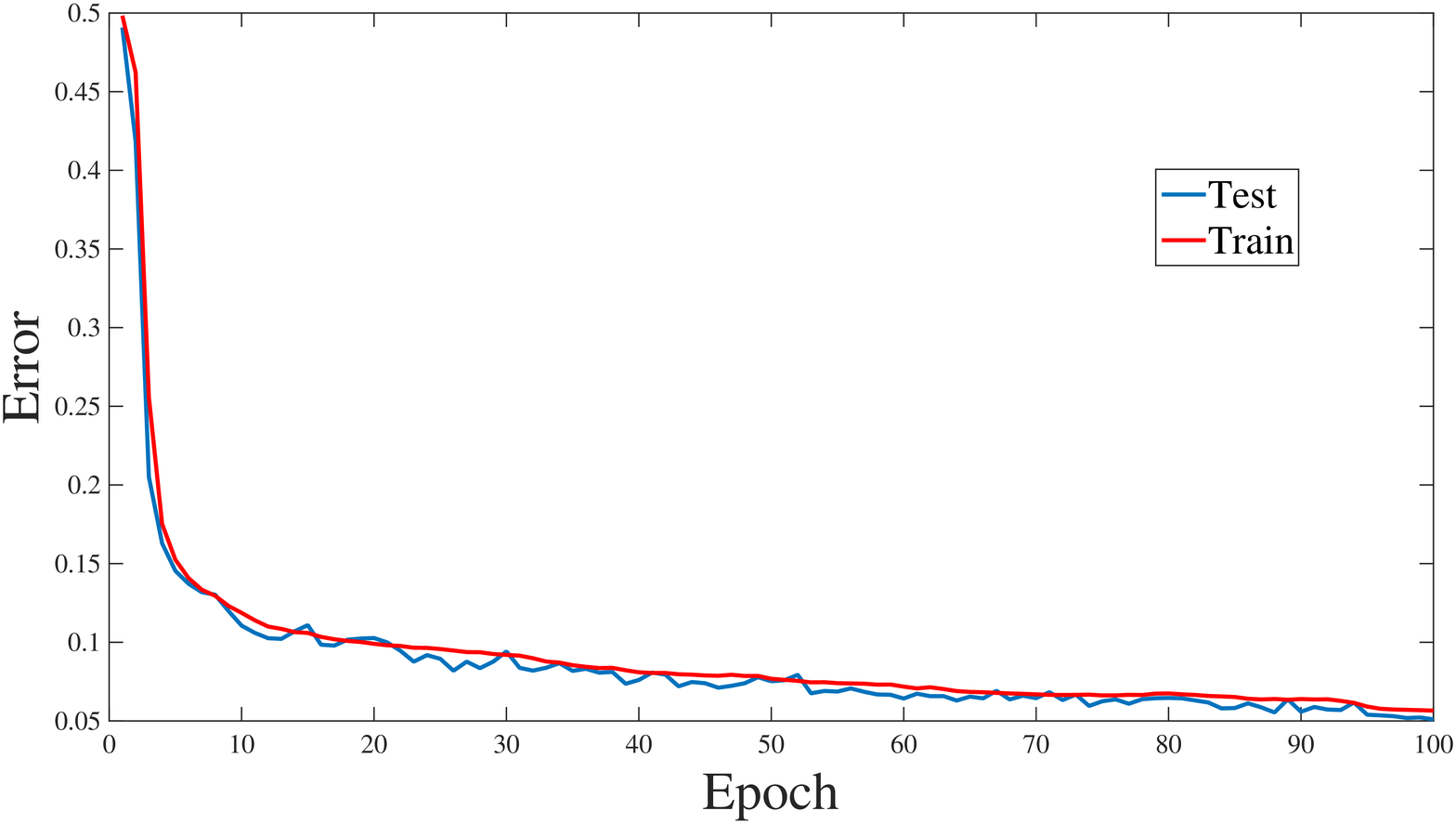}
\end{subfigure}
{\caption{Contours extracted from the MPEG7 Database and the error plot for training.}
\label{Training}}
\end{figure}
\subsection{Building Representative Datasets and Implementation}
In order to train for invariance, we need to build a dataset with two major attributes: 
First, it needs to contain a large number of examples of the transformation and second, 
 the curves involved in the training need to have sufficient richness in terms of different 
 patterns of sharp edges,  corners, smoothness, noise and sampling factors to ensure sufficient
 generalizability of the model. 
To sufficiently span the space of Euclidean transformations, we generate random two dimensional 
 rotations by uniformly sampling angles from $[-\pi,\pi]$. 
The curves are normalized by removing the mean and dividing by the standard deviation thereby 
 achieving invariance to translations and uniform scaling. 
The contours are extracted from the shapes of the MPEG7 Database (\cite{latecki2000shape}) as 
 shown in first part of Figure \ref{Training}. 
It comprises a total of $1400$ shapes containing $70$ different categories of objects. 
$700$ of the total were used for training and $350$ each for testing and validation. 
The positive examples are constructed by taking a curve and randomly transforming it by a rotation, 
 translation and reflection and pairing them together. 
The negative examples are obtained by pairing curves which are deemed dissimilar as explained in 
 Section \ref{MultiScale}. 
The contours are extracted and each contour is 
 sub-sampled to $500$ points. 
We build the training dataset of $10,000$ examples with approximately $50\%$ each for the positive 
 and negative examples.  
The network and training is performed using the Torch library \cite{collobert2002torch}. 
We trained using Adagrad \cite{duchi2011adaptive} at a learning rate of  $5\times 10^{-4}$ and a batch size of $10$. We set the contrastive loss hyperparameter {\it margin} $\mu = 1$ and Figure \ref{Training} shows the error plot
for training and the convergence of the loss to a minimum. The rest of this work describes how we can observe and extend the efficacy of the trained network on new data.

\section{Multi-Scale Representations}
\label{MultiScale}
Invariant representations at varying levels of abstraction have a theoretical interest as well as 
 practical importance to them. 
Enumeration at different scales enables a hierarchical method of analysis which is useful when
 there is noise and hence stability is desired in the invariant. 
As mentioned in Section \ref{Background}, the invariants constructed from scale-space methods 
 and integral invariants, naturally allow for such a decomposition by construction. 

\begin{figure}[t]
\begin{floatrow}
\ffigbox[1.3\FBwidth][]
{
\begin{tikzpicture}[every node/.style={scale=0.35}]
\node(curve1)[curvebox]{\includegraphics[scale = 0.25]
{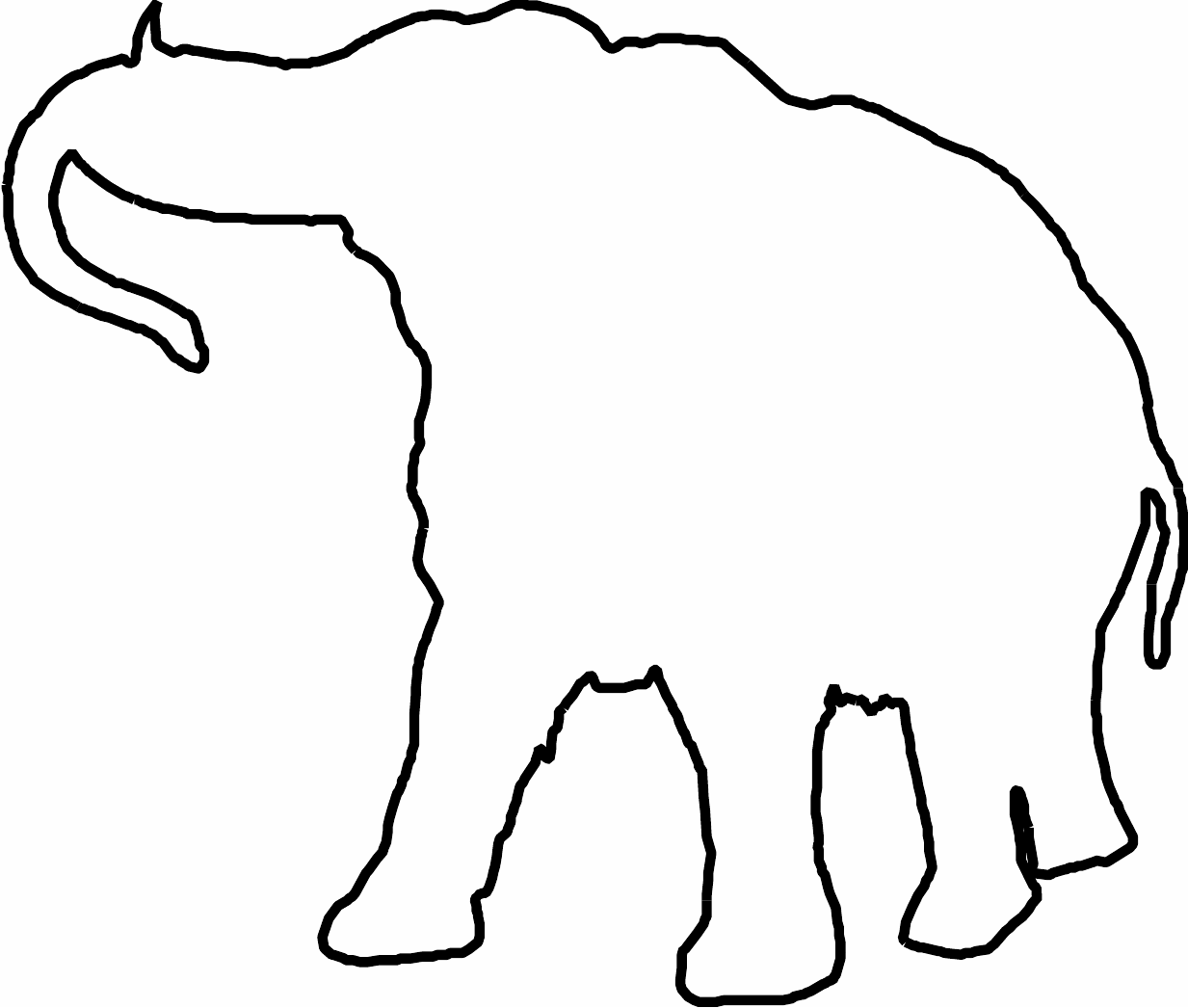}};

\node(curve2)[curvebox,below of = curve1,yshift=-2.5cm]{\includegraphics[scale = 0.25]
{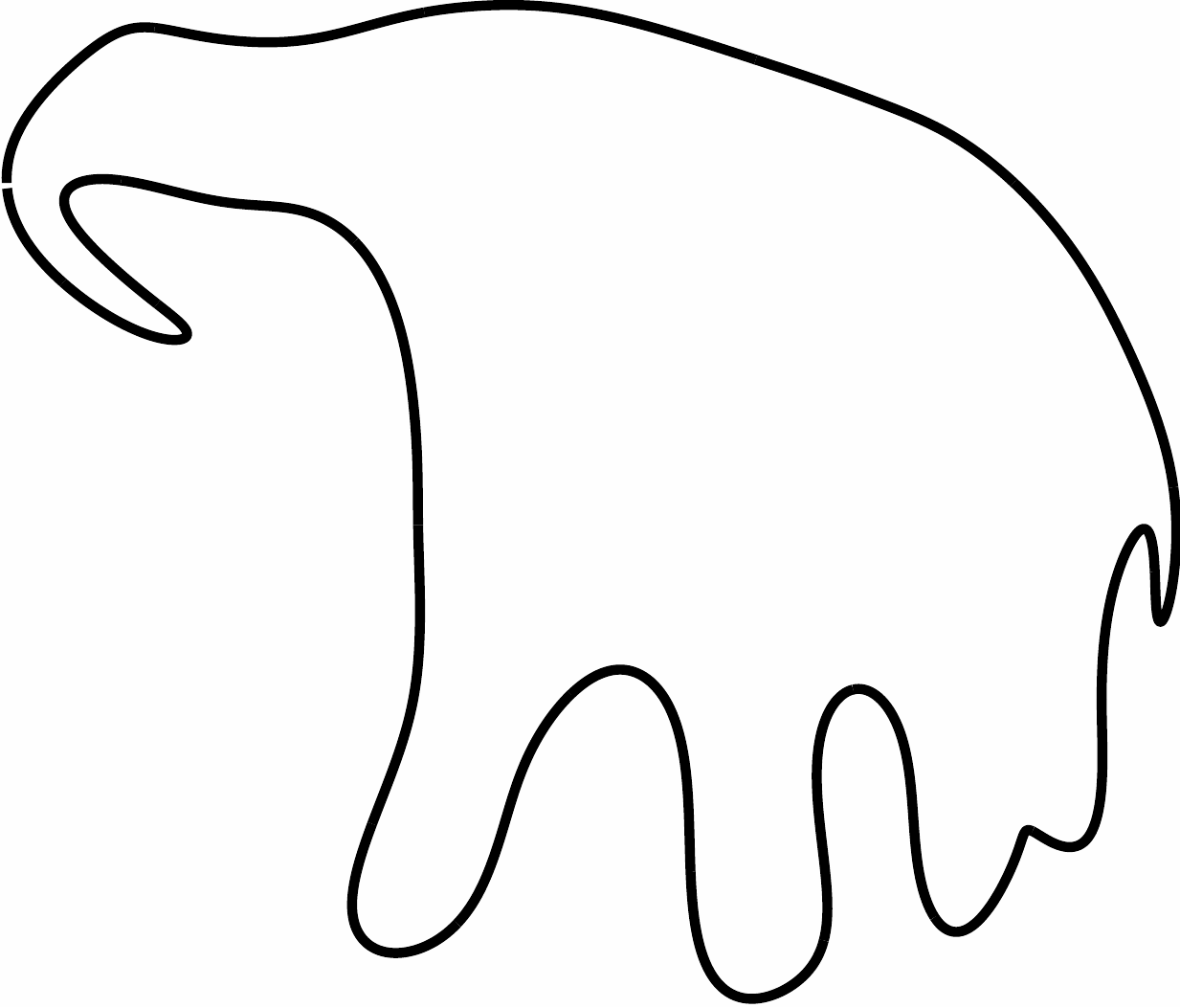}};

\node(curve3)[curvebox,below of = curve2,yshift=-2.5cm]{\includegraphics[scale = 0.225]
{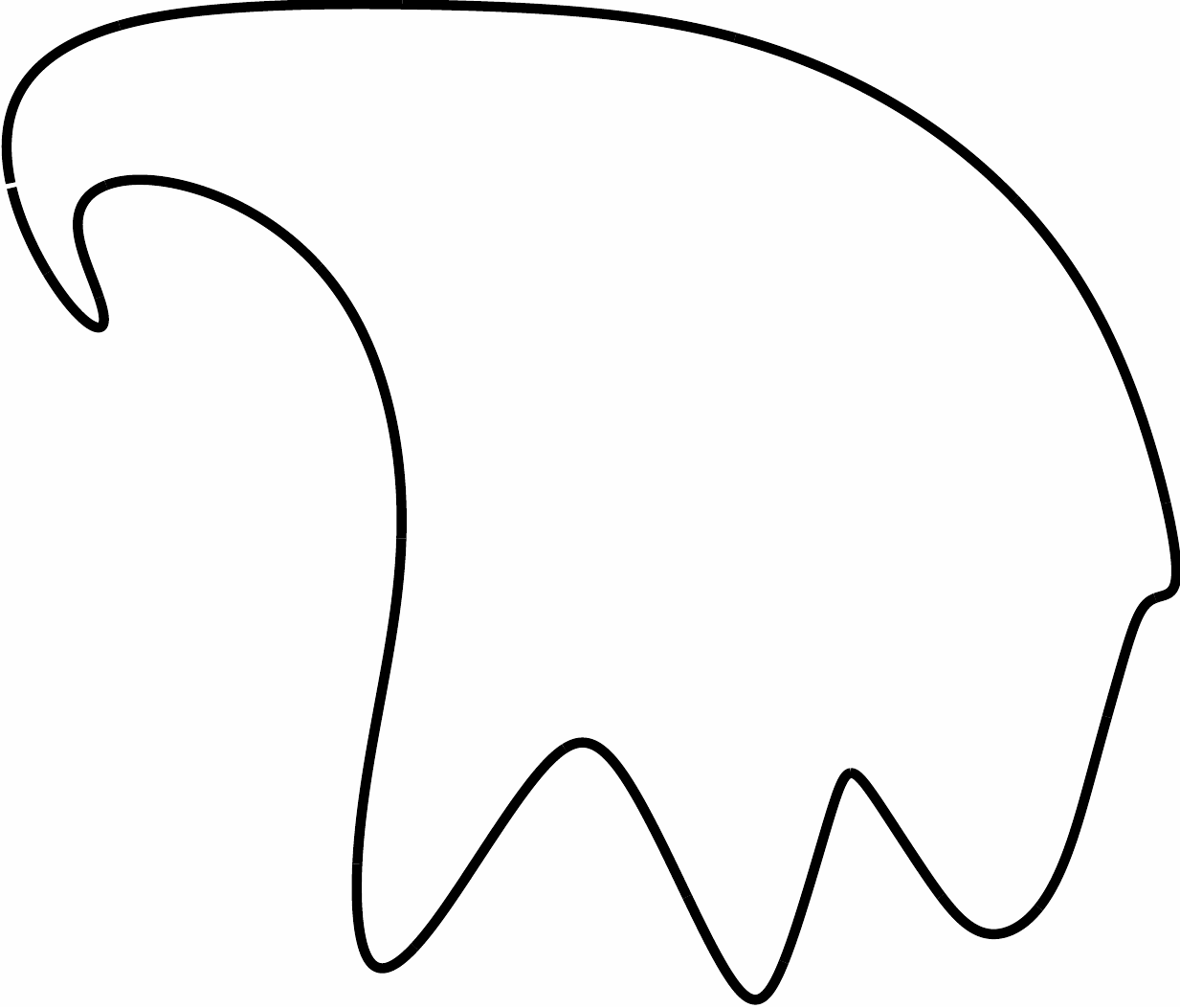}};

\node(curve4)[curvebox,below of = curve3,yshift=-2.5cm]{\includegraphics[scale = 0.22]
{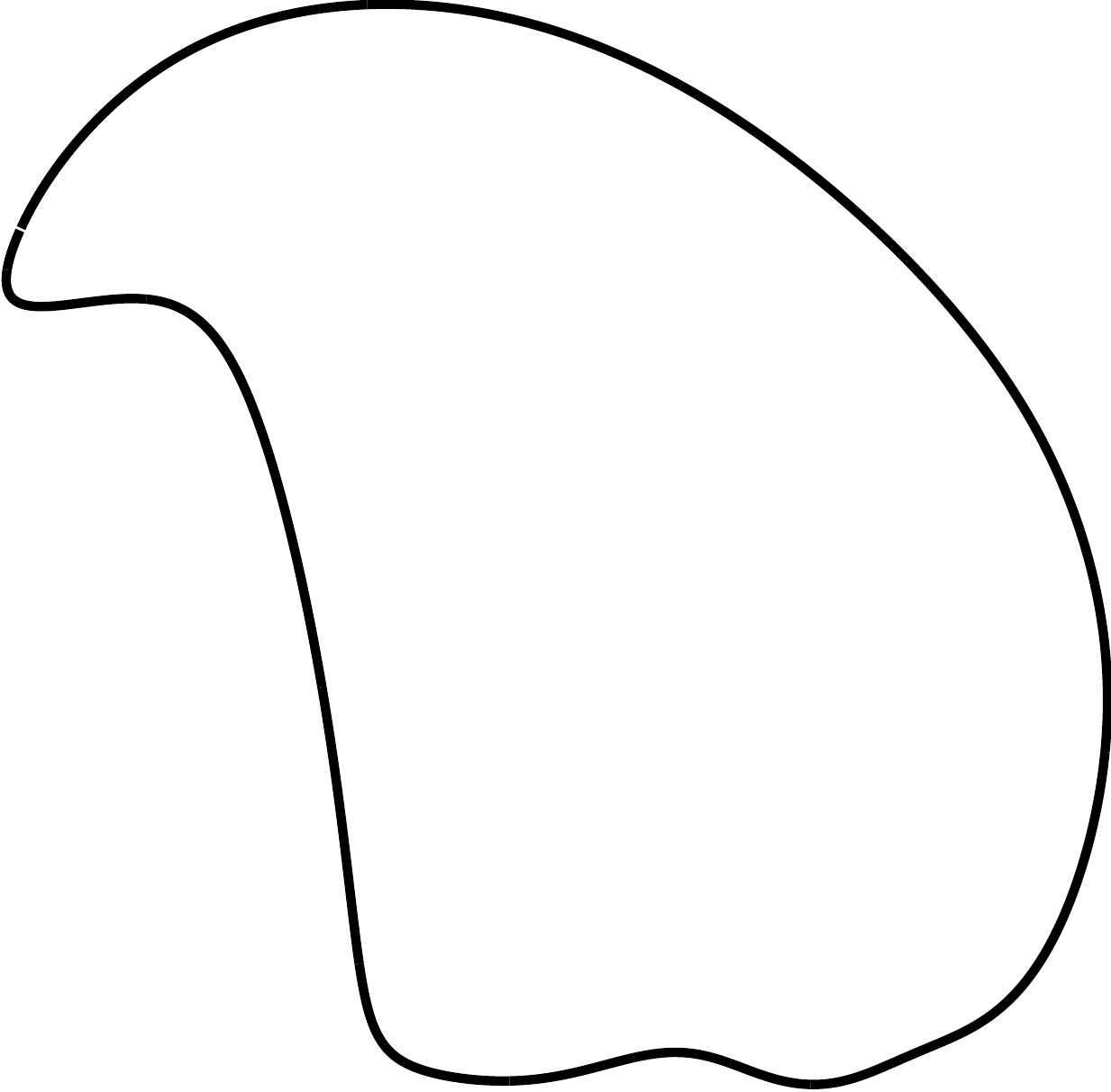}};

\node(curve5)[curvebox,below of = curve4,yshift=-2.5cm]{\includegraphics[scale = 0.08]
{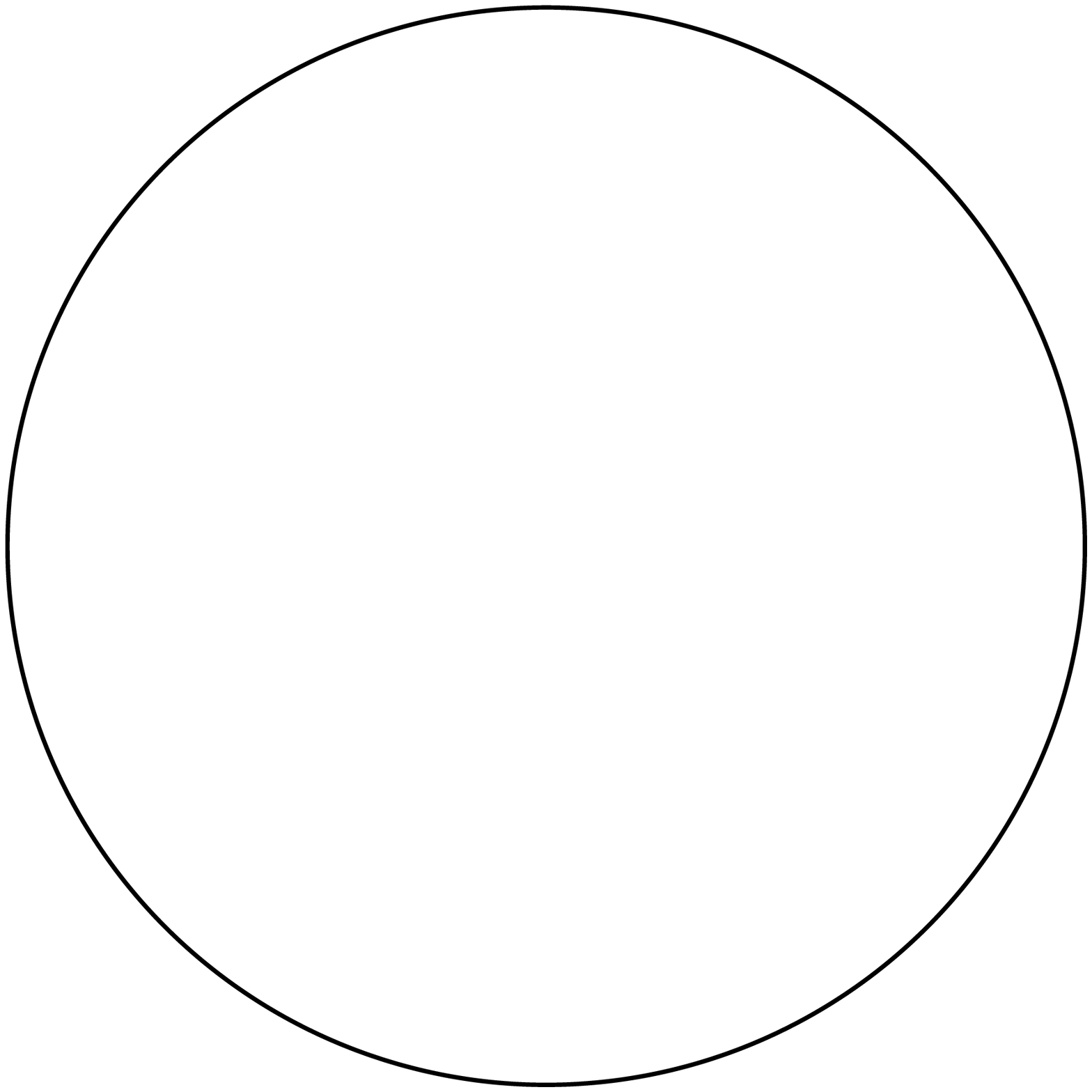}};

\node(sig1)[curvebox,right of = curve1,xshift=6cm]{\includegraphics[scale = 0.25]
{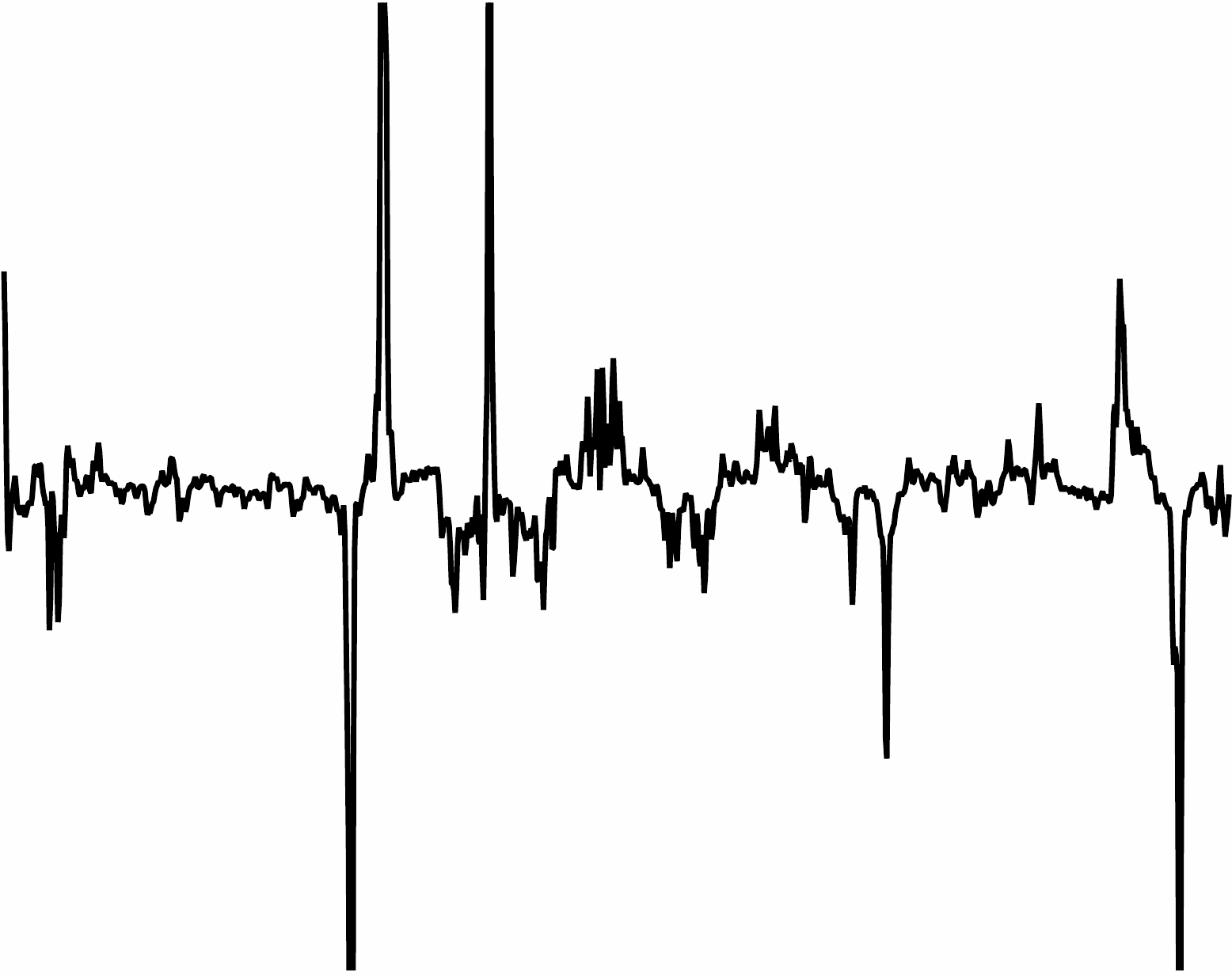}};

\node(sig2)[curvebox,right of = curve2,xshift=6cm]{\includegraphics[scale = 0.25]
{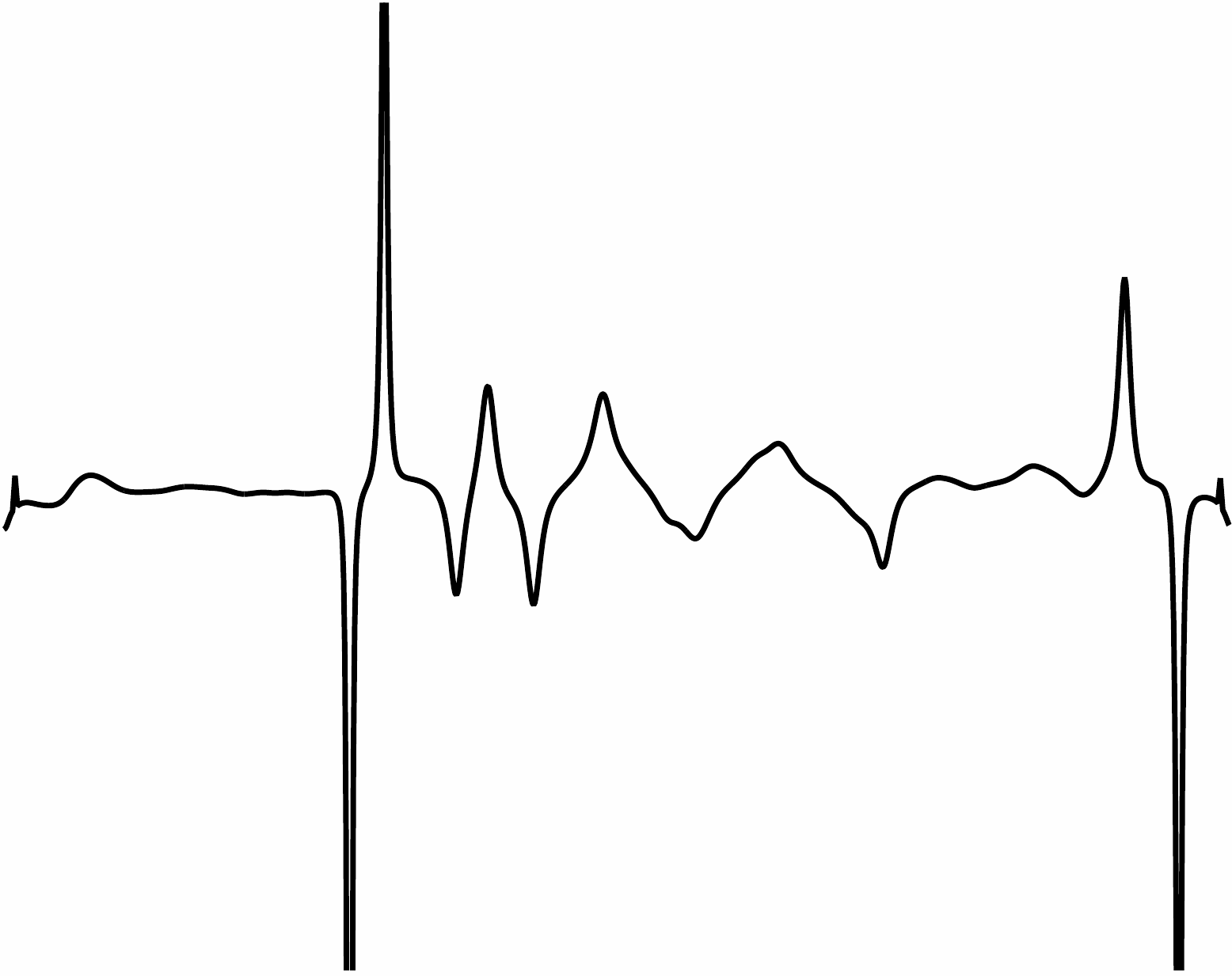}};

\node(sig3)[curvebox,right of = curve3,xshift=6cm]{\includegraphics[scale = 0.25]
{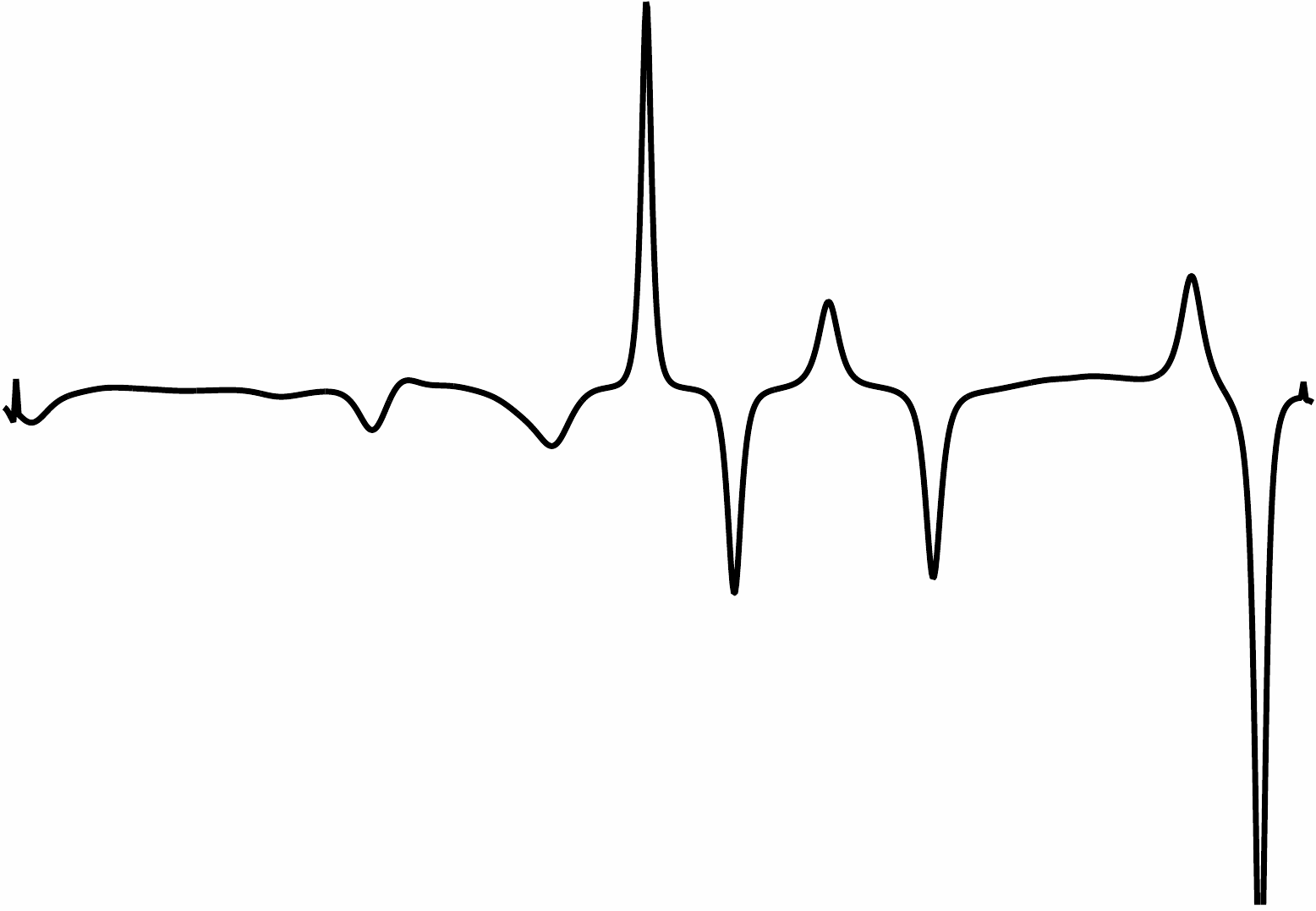}};

\node(sig4)[curvebox,right of = curve4,xshift=6cm]{\includegraphics[scale = 0.25]
{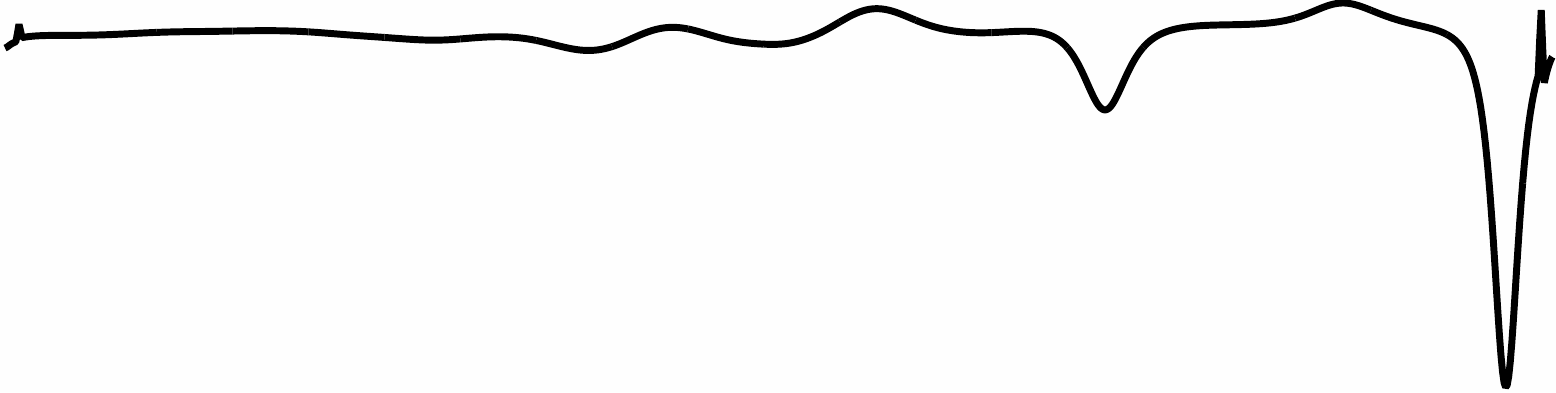}};

\node(dot1)[textbox,below of = curve4]{\Big{.}};

\node(sig5)[curvebox,right of = curve5,xshift=6cm]{\includegraphics[scale = 0.12]
{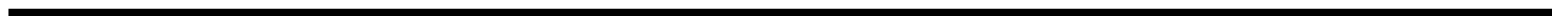}};
\draw [loosely dotted] (curve4) -- (curve5);
\node(curveText)[textbox,above of = sig1, yshift = 1.5cm]{\Huge{Curvature:  $\kappa$}};
\end{tikzpicture}
}{\caption{Curve evolution and the corresponding curvature profile.} 
\label{curvature_flow}}
\capbtabbox[1.2\FBwidth][]
{
\resizebox{0.485\textwidth}{!}{
\def\arraystretch{8}
\setlength\tabcolsep{0.001\textwidth}
\begin{tabular}{|c|r p{3cm}|c|}
\hline 
\rule{0pt}{1pt} \LARGE{Positive Example} & \multicolumn{2}{|c|}{\LARGE{Negative} Example}& \LARGE{Scale Index} \\
\hline
\multirow{4}{*}{
\begin{tikzpicture}
\hspace{-0.4cm}\node(curveA)[curvebox]{\includegraphics[scale=0.08] {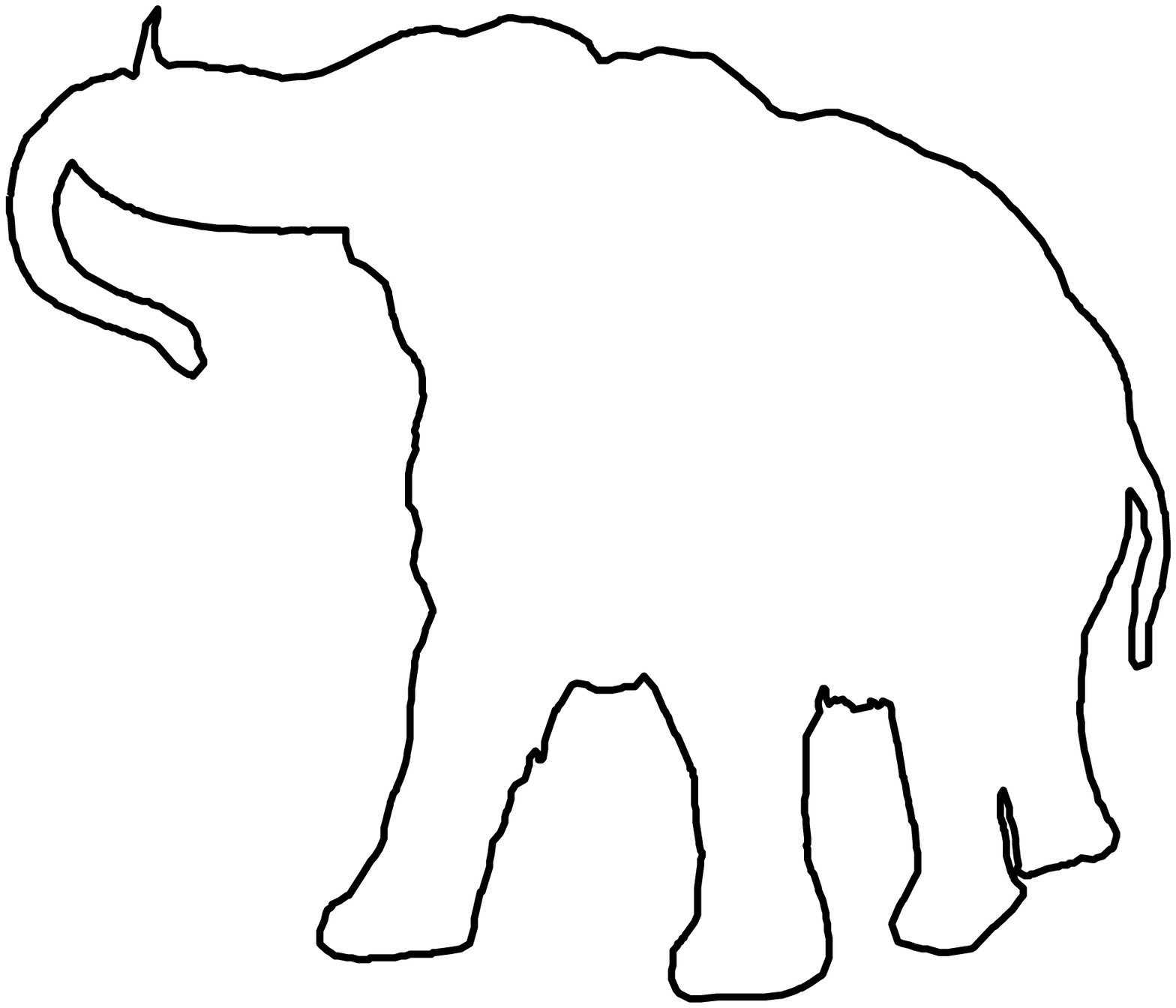}};
\node(curveB)[curvebox, right of = curveA, xshift = 0.120\textwidth]{\includegraphics[scale=0.08] {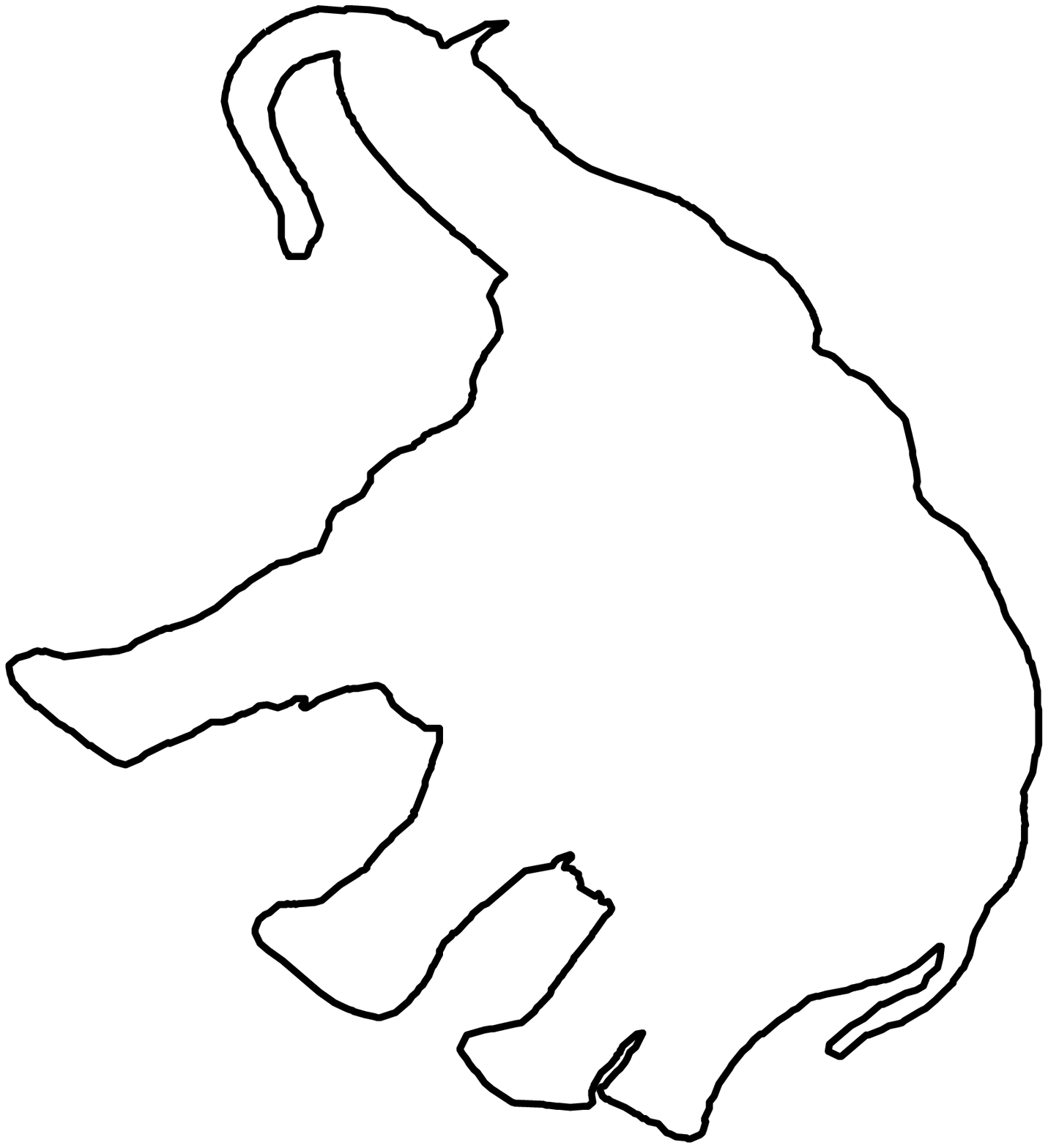}};
\end{tikzpicture}}
&\includegraphics[scale=0.08]{Figures/fig_main.eps}&\includegraphics[scale=0.12]{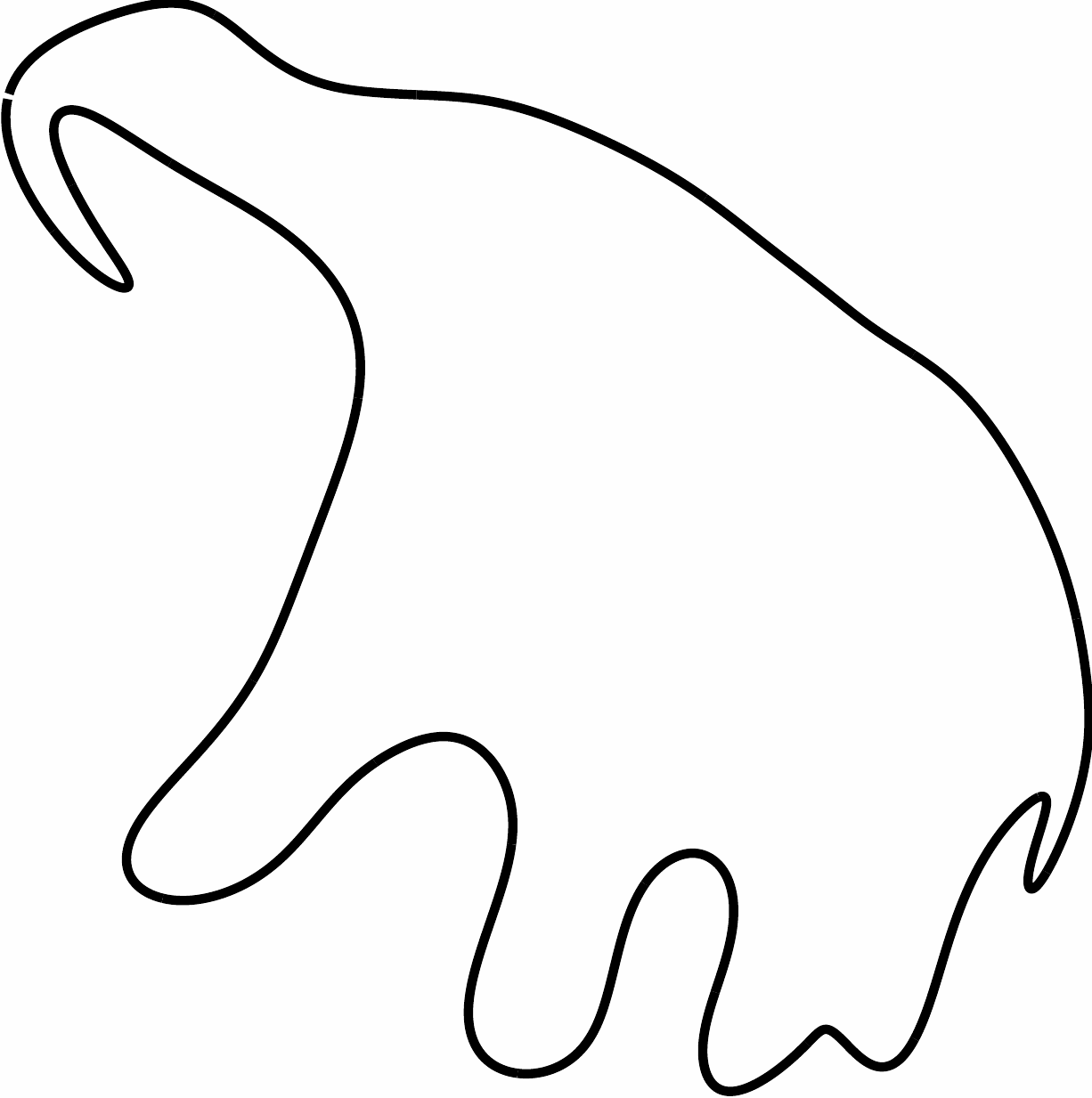}&
\multirow{4}{*}{
\begin{tikzpicture}
\node(scaleTextTop)[textbox] {\Huge{Low}};
\node(scaleTextBottom)[textbox, below of = scaleTextTop, yshift = -7cm]{\Huge{High}};
\draw [decoration={markings,mark=at position 1 with
    {\arrow[scale=3,>=stealth]{>}}},postaction={decorate}] (scaleTextTop) -- (scaleTextBottom) node [midway, right=0.5cm] {};    
\end{tikzpicture}
}\\
\cline{2-3}
&\hspace{0.1cm}\includegraphics[scale=0.08]{Figures/fig_main.eps}&\includegraphics[scale=0.11]{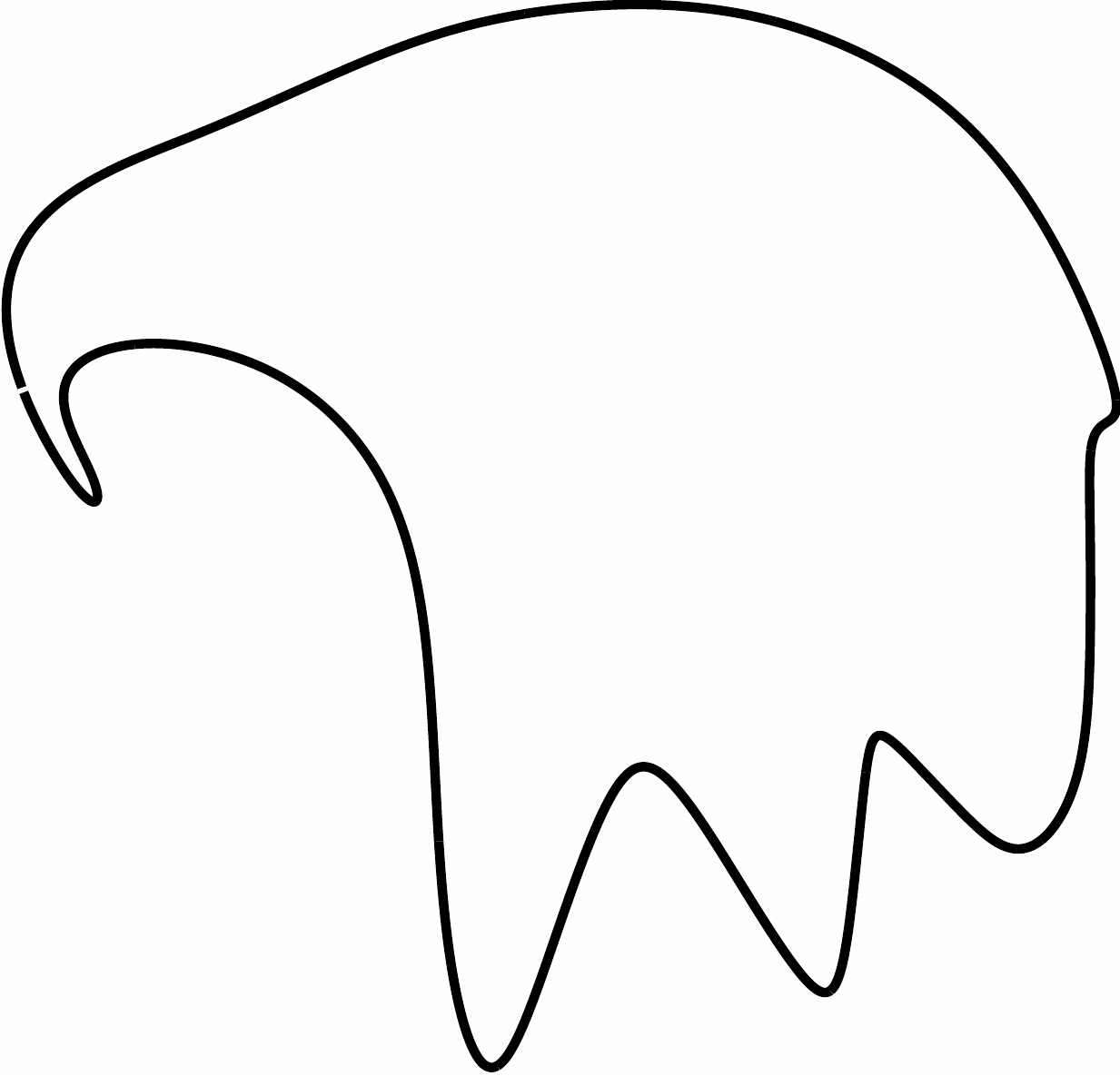}&\\ \cline{2-3}
&\includegraphics[scale=0.08]{Figures/fig_main.eps}&\includegraphics[scale=0.11]{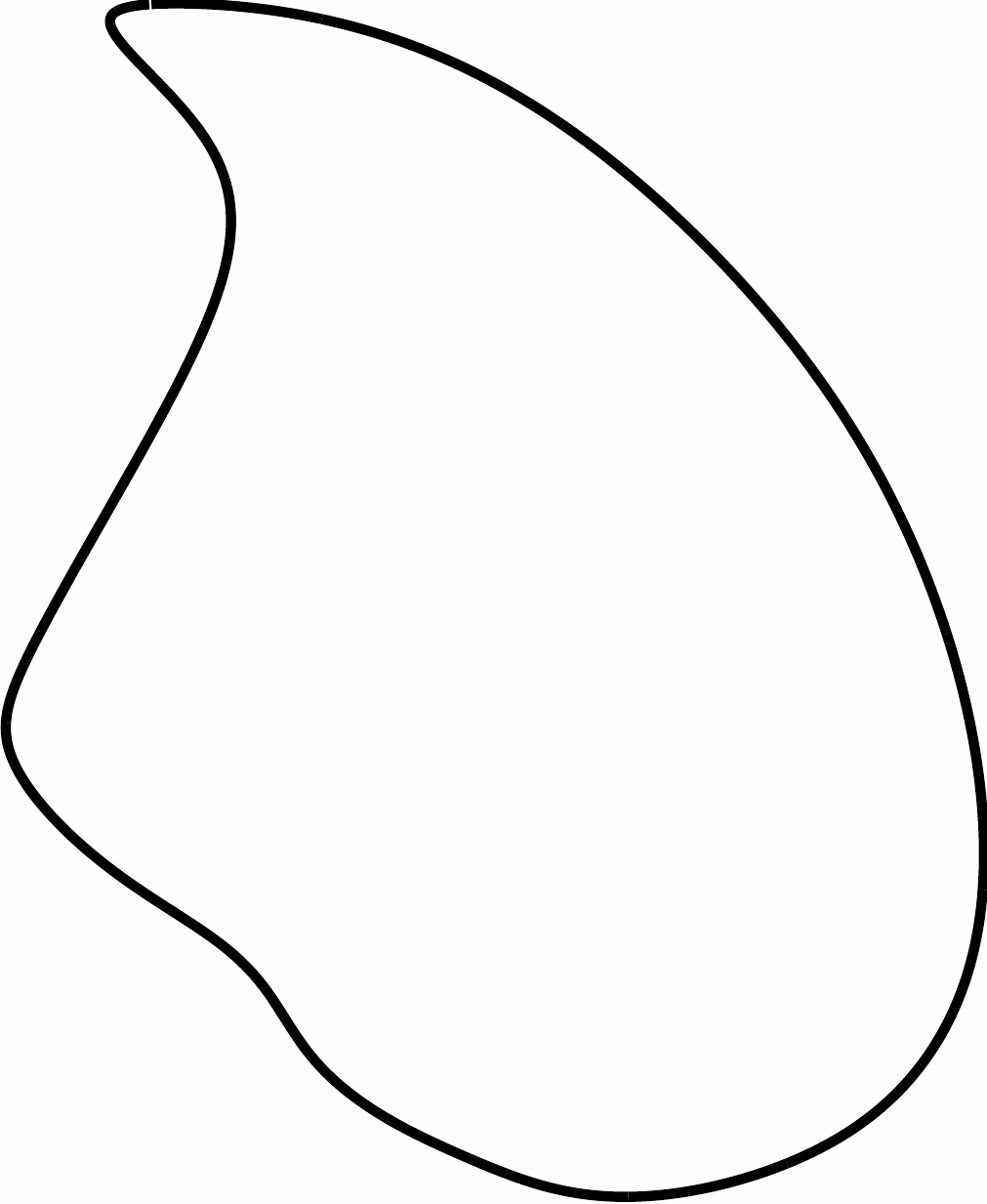}&\\ \cline{2-3}
&\includegraphics[scale=0.08]{Figures/fig_main.eps}&\includegraphics[scale=0.05]{Figures/fig5.eps}&\\
\hline
\end{tabular}
}
}{\caption{Examples of training pairs for different scales. 
           Each row indicates the pattern of training examples for a different scale.} 
           \label{table_train}
\label{training_pairs}}
\end{floatrow}
\end{figure}

A valuable insight for multi-scale representations is provided in the theorems of Gage, Hamilton 
 and Grayson (\cite{gage1986heat}; \cite{grayson1987heat}). 
It says that if we evolve any smooth non-intersecting planar curve with mean curvature flow, 
 which is invariant to Euclidean transformations, it will ultimately converge  into a circle before 
  vanishing into a point. 
The curvature corresponding to this evolution follows a profile as shown in Figure \ref{curvature_flow}, 
 going from a possibly noisy descriptive feature to a constant function. 
In our framework, we observe an analogous behavior in a data-dependent setting. 
The positive part of the loss function $(\lambda =1)$ forces the network to push the outputs 
 of the positive examples closer, whereas the negative part $(\lambda = 0)$ 
 forces the weights of network to push the outputs of the negative examples apart, 
 beyond the distance barrier of $\mu$. 
If the training data does not contain any negative example, it is easy to see that the weights
 of the network will converge to a point which will yield a constant output that trivially minimizes 
 the loss function in Equation \ref{siamese_loss}.  
This is analogous to that point in curvature flow which yields a circle and therefore has
 a constant curvature. 

\begin{figure}[t]
\hspace{-0.35\textwidth}
\begin{subfigure}[h]{0.2\textwidth}
\includegraphics[scale=0.19]{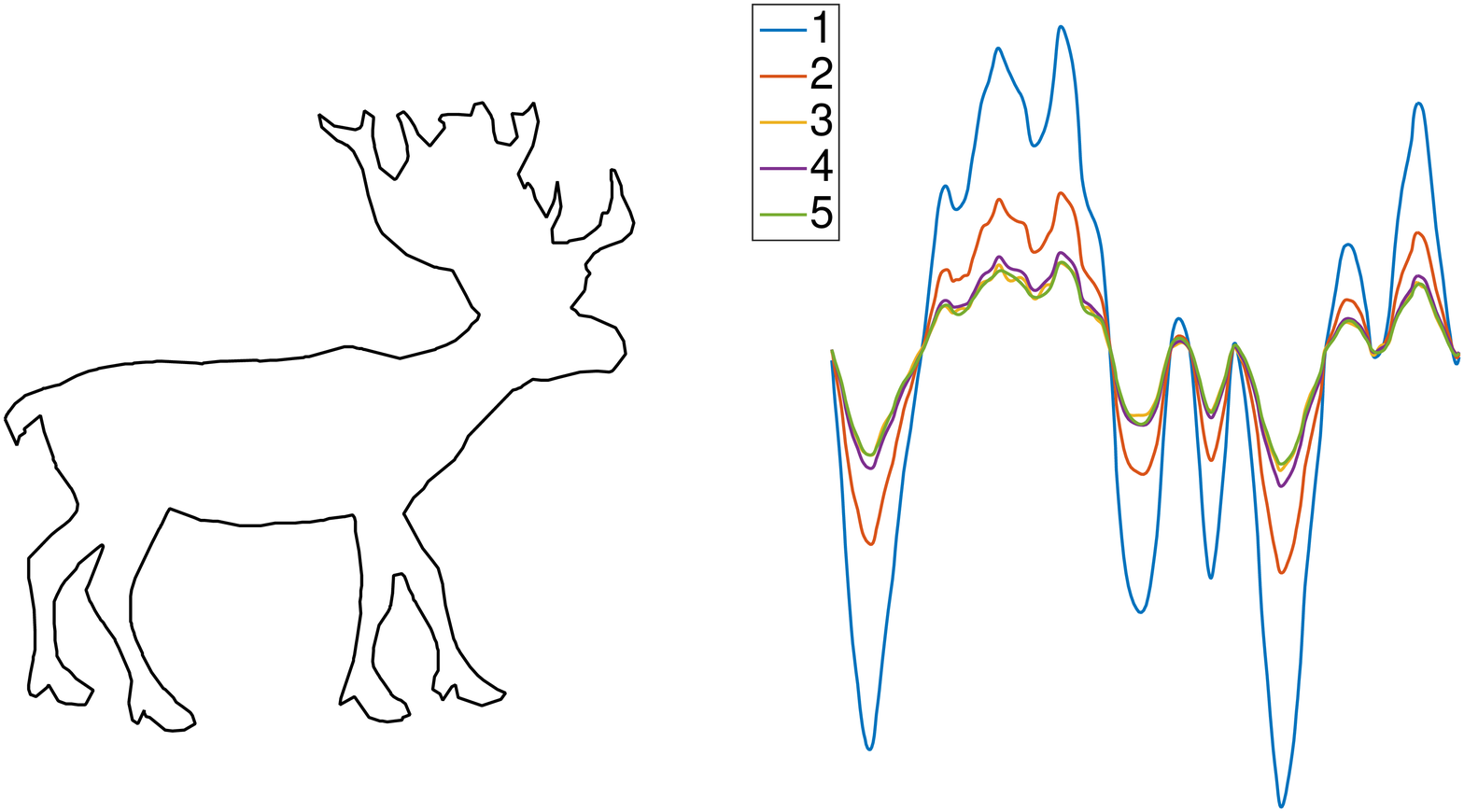}
\end{subfigure}
\hspace{0.35\textwidth}
\begin{subfigure}[h]{0.3\textwidth}
\includegraphics[scale=0.19]{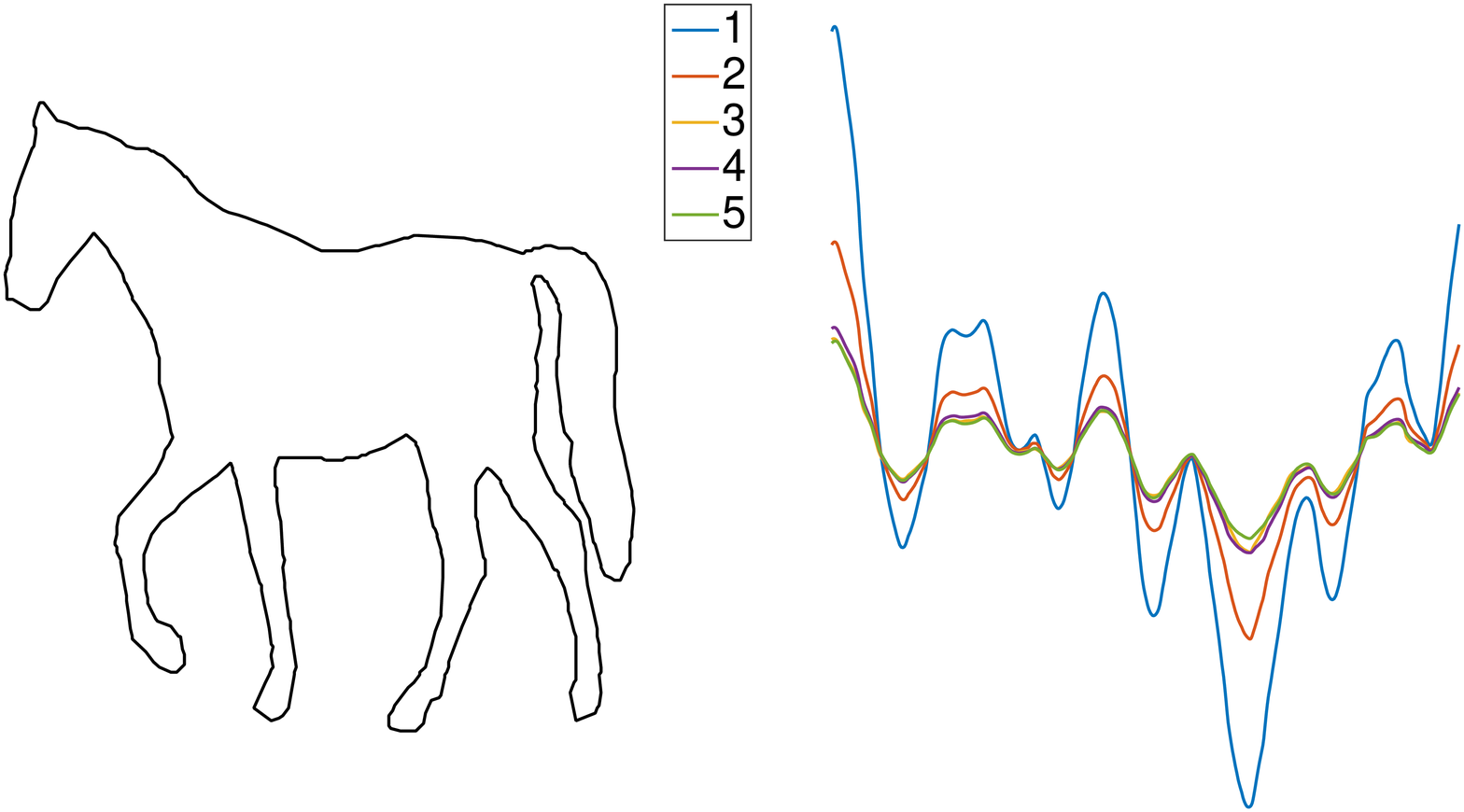}
\end{subfigure}
\caption{Experiments with multi-scale representations. 
  Each signature is the output of a network trained on a dataset with training examples 
  formed as per the rows of Table \ref{training_pairs}. 
  Index1 indicates low and $5$ indicates a higher level of abstraction.}
\label{Multiscale_Example}
\end{figure}

Designing the negative examples of the training data provides the means to obtain
 a multi-scale representation. Since we are training for a \emph{local} descriptor of a curve, that is, a function whose value at a point depends only on its local neighborhood, a negative example must pair curves such that corresponding points on each curve must have \emph{different local} neighborhoods.
One such possibility is to construct negative examples which pair curves with their smoothed or 
 evolved versions as in Table \ref{training_pairs}.  
Minimizing the loss function in equation \ref{siamese_loss} would lead to an action which pushes apart
 the signatures of the curve and its evolved or smoothed counterpart, thereby injecting the signature
 with fidelity and descriptiveness. 
We construct separate data-sets where the negative examples are drawn as shown in the rows of 
 Table\ref{training_pairs} and train a network model for each of them using the loss function
 \ref{siamese_loss}. 
In our experiments we perform smoothing by using a local polynomial regression with weighted linear
 least squares for obtaining the evolved contour. 
Figure \ref{Multiscale_Example} shows the outputs of these different networks which demonstrate
 a scale-space like behavior. 
\section{Experiments and Discussion}
\label{Experiments}
\begin{figure}[t]
\hspace{-0.65\textwidth}
\begin{subfigure}[h]{0.2\textwidth}
\centering
\includegraphics[scale=0.3]{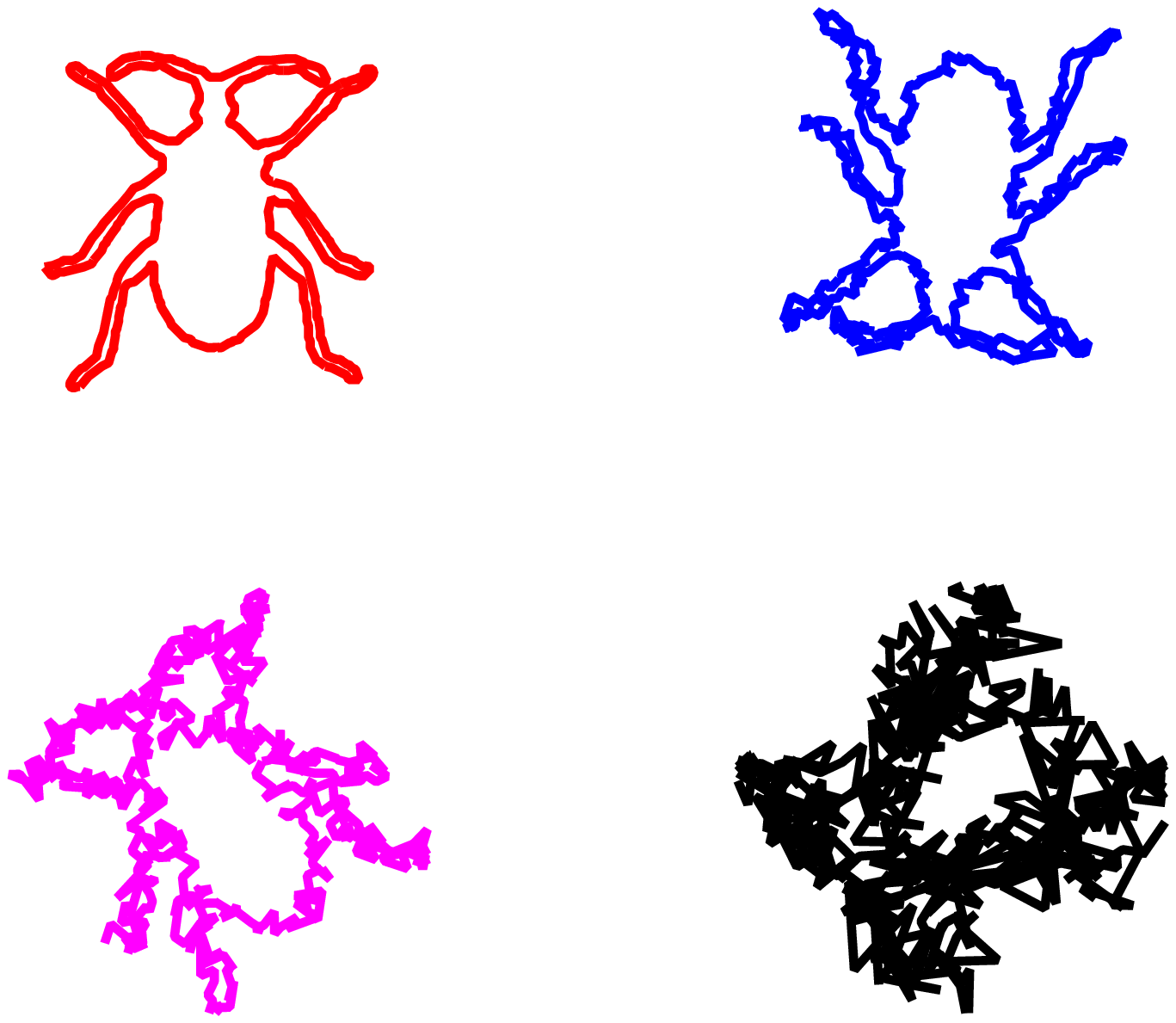}
\end{subfigure}
\hspace{0.15\textwidth}
\begin{subfigure}[h]{0.2\textwidth}
\centering
\includegraphics[scale=0.25]{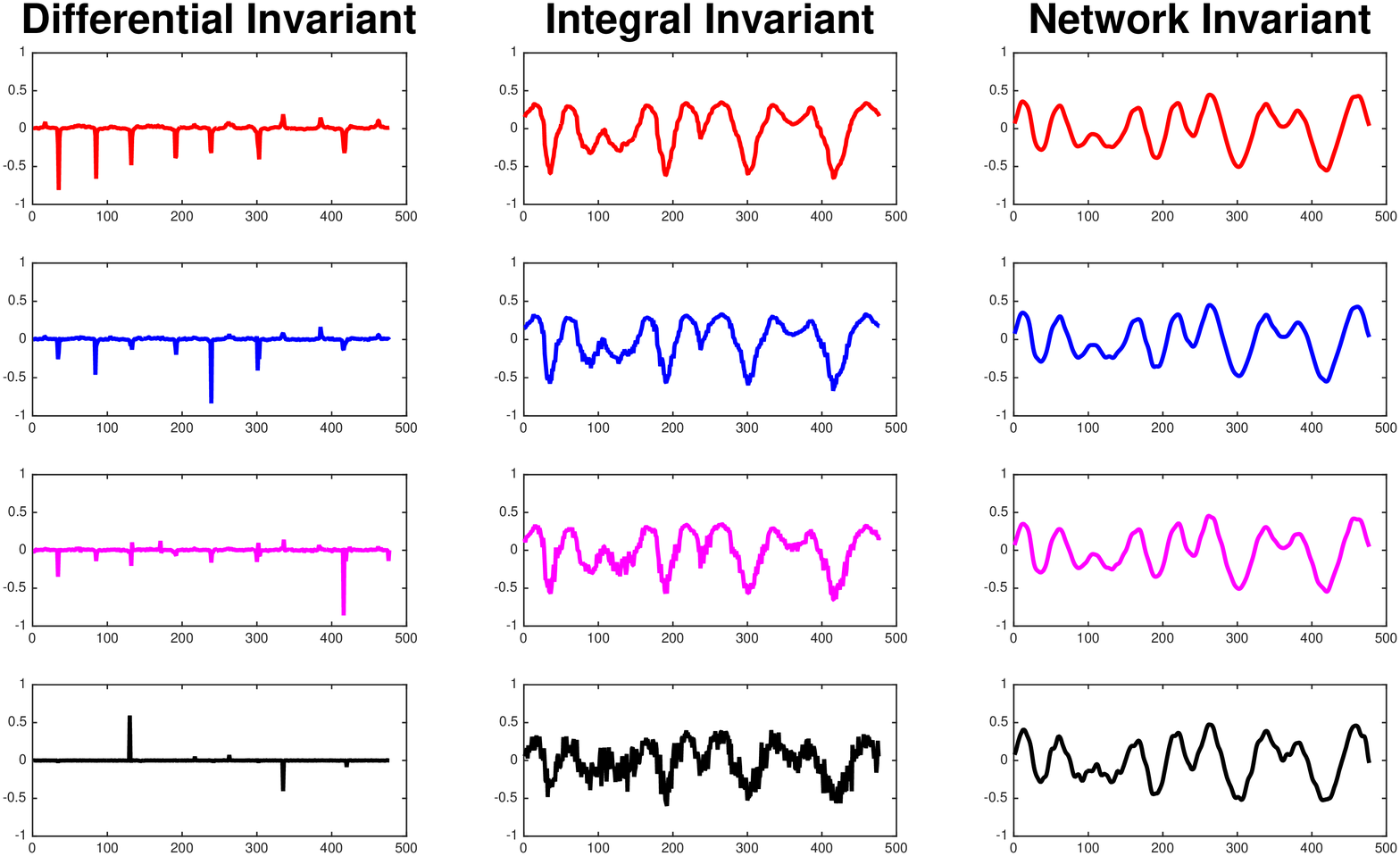}
\end{subfigure}

\hspace{-0.65\textwidth}
\begin{subfigure}[h]{0.2\textwidth}
\centering
\includegraphics[scale=0.3]{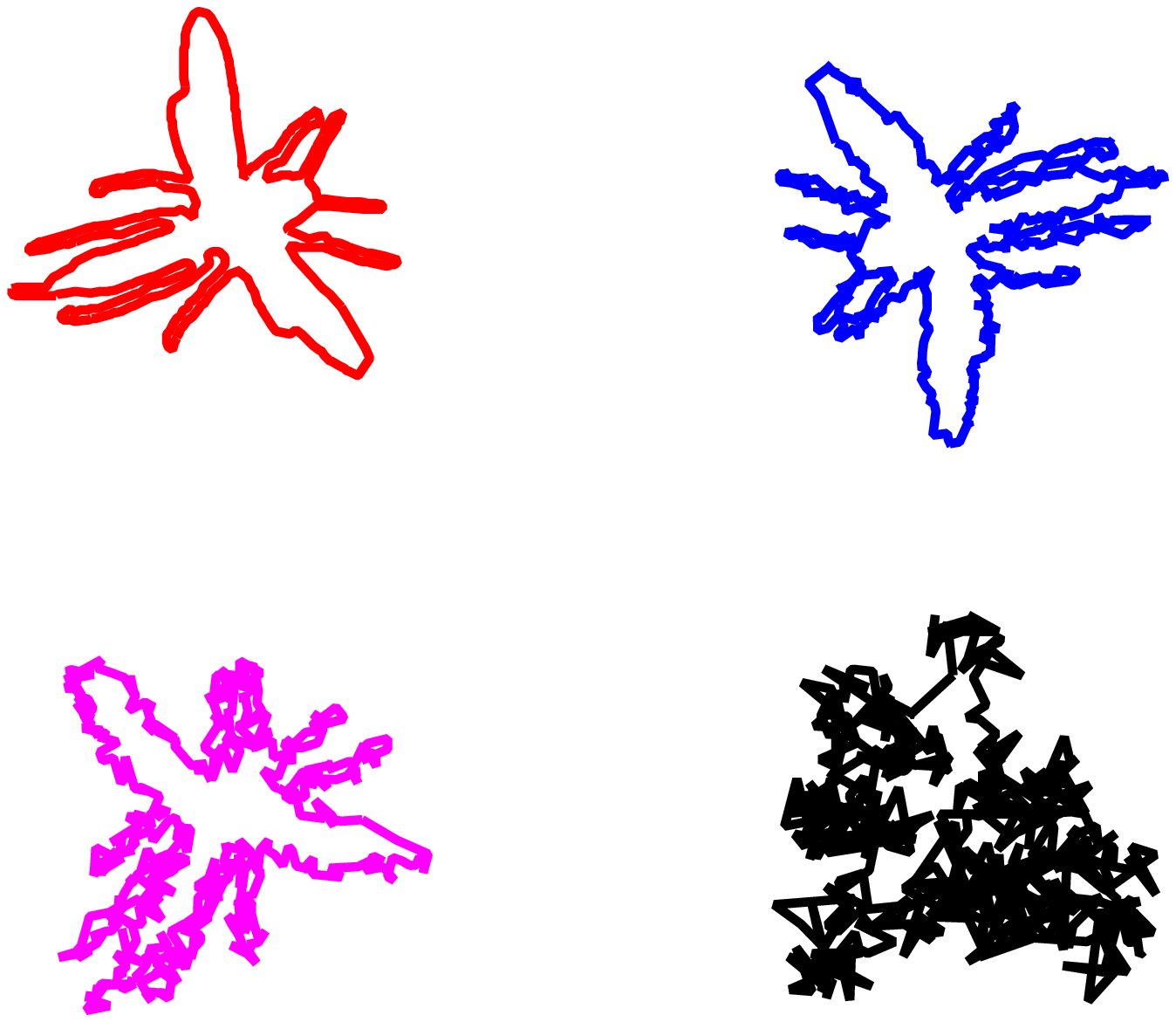}
\end{subfigure}
\hspace{0.15\textwidth}
\begin{subfigure}[h]{0.2\textwidth}
\centering
\includegraphics[scale=0.25]{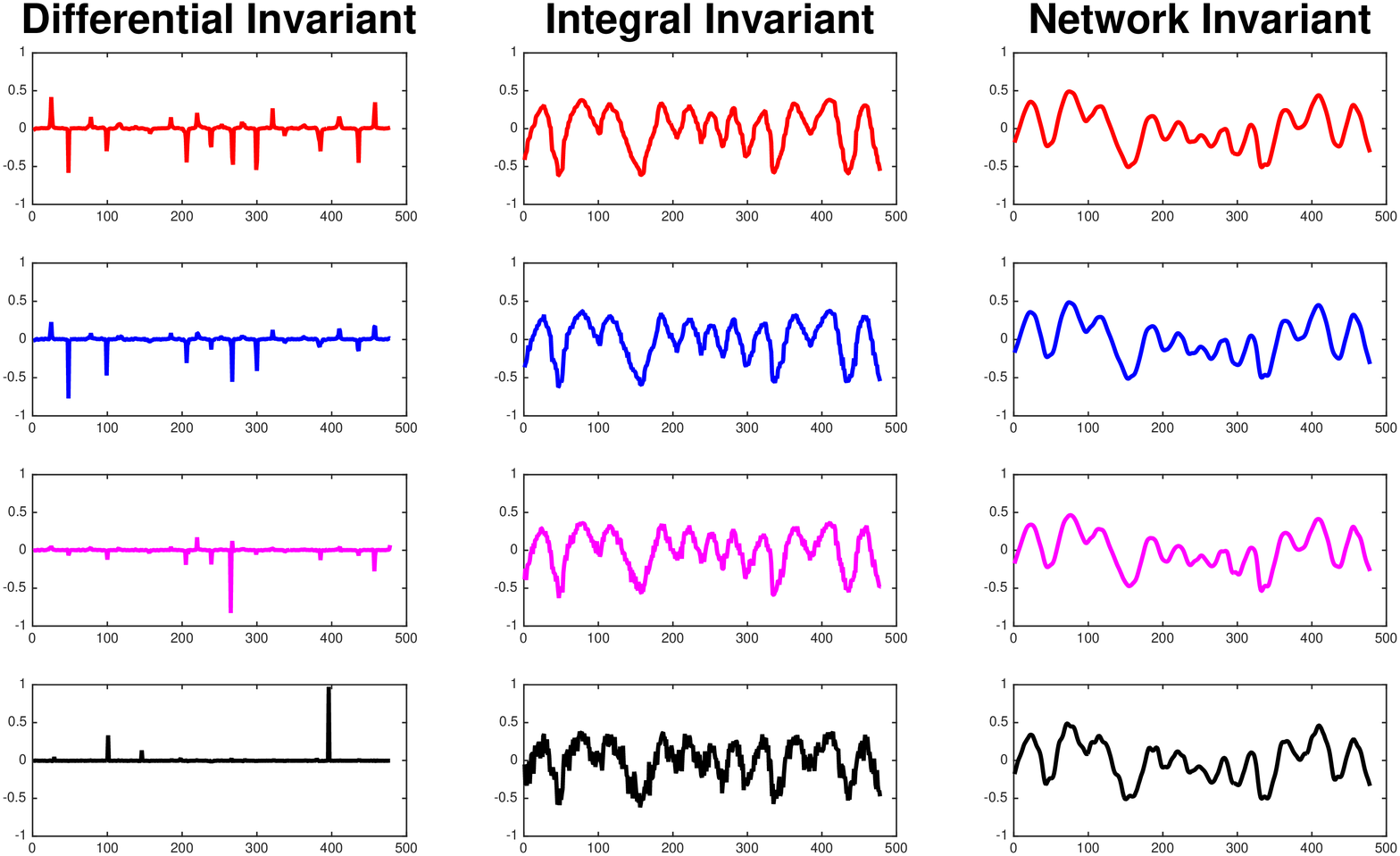}
\end{subfigure}
\caption{Stability of different signatures in varying levels noise and Euclidean transformations. 
              The correspondence for the shape and the signature is the color. 
              All signatures are normalized.}
\label{Signatures_Noise}
\end{figure}

\begin{figure}[t]
\hspace{-0.4\textwidth}
\begin{subfigure}[h]{0.2\textwidth}
\centering
\includegraphics[scale=0.18]{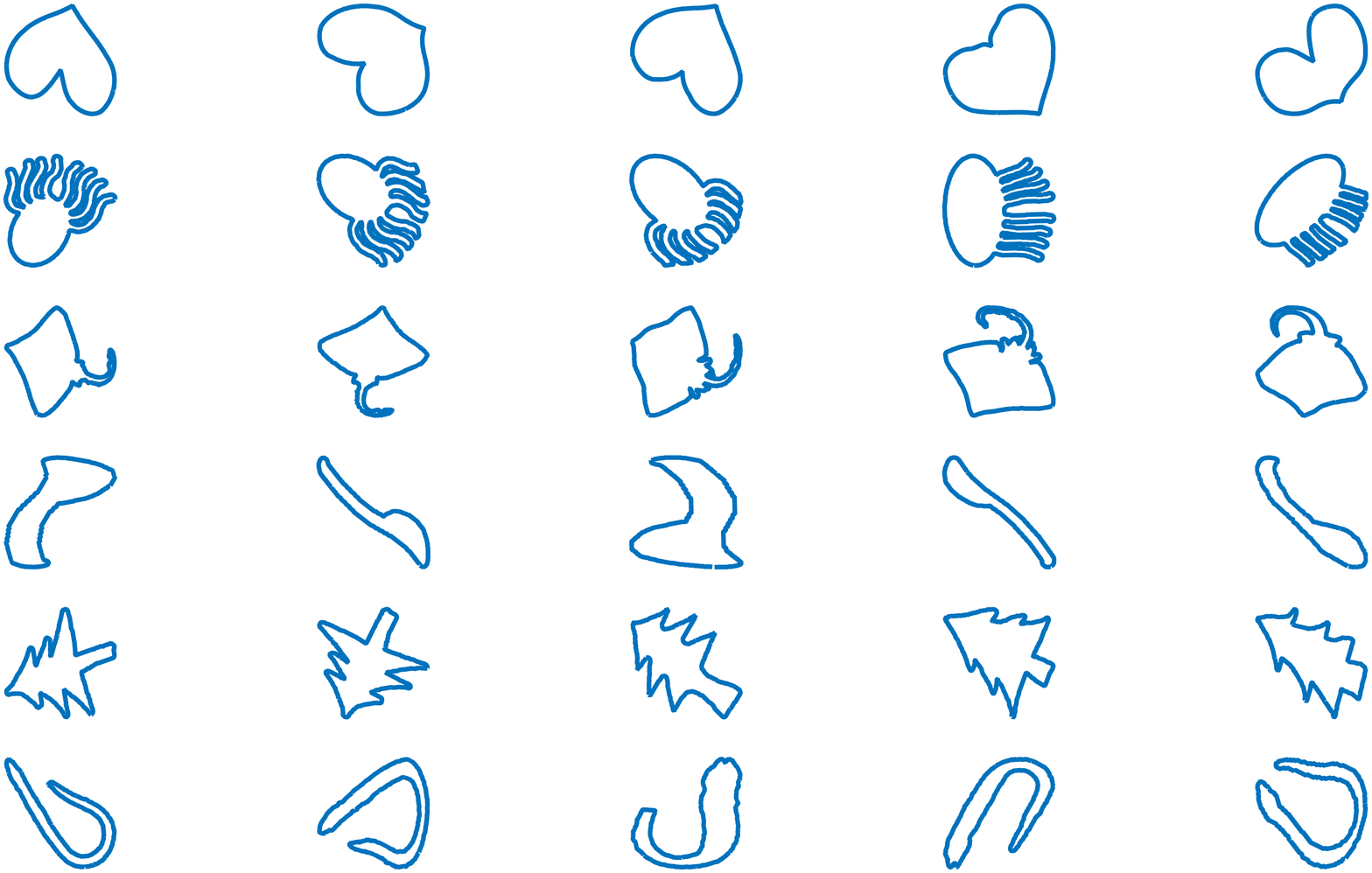}
\end{subfigure}
\hspace{0.31\textwidth}
\begin{subfigure}[h]{0.3\textwidth}
\centering
\includegraphics[scale=0.2]{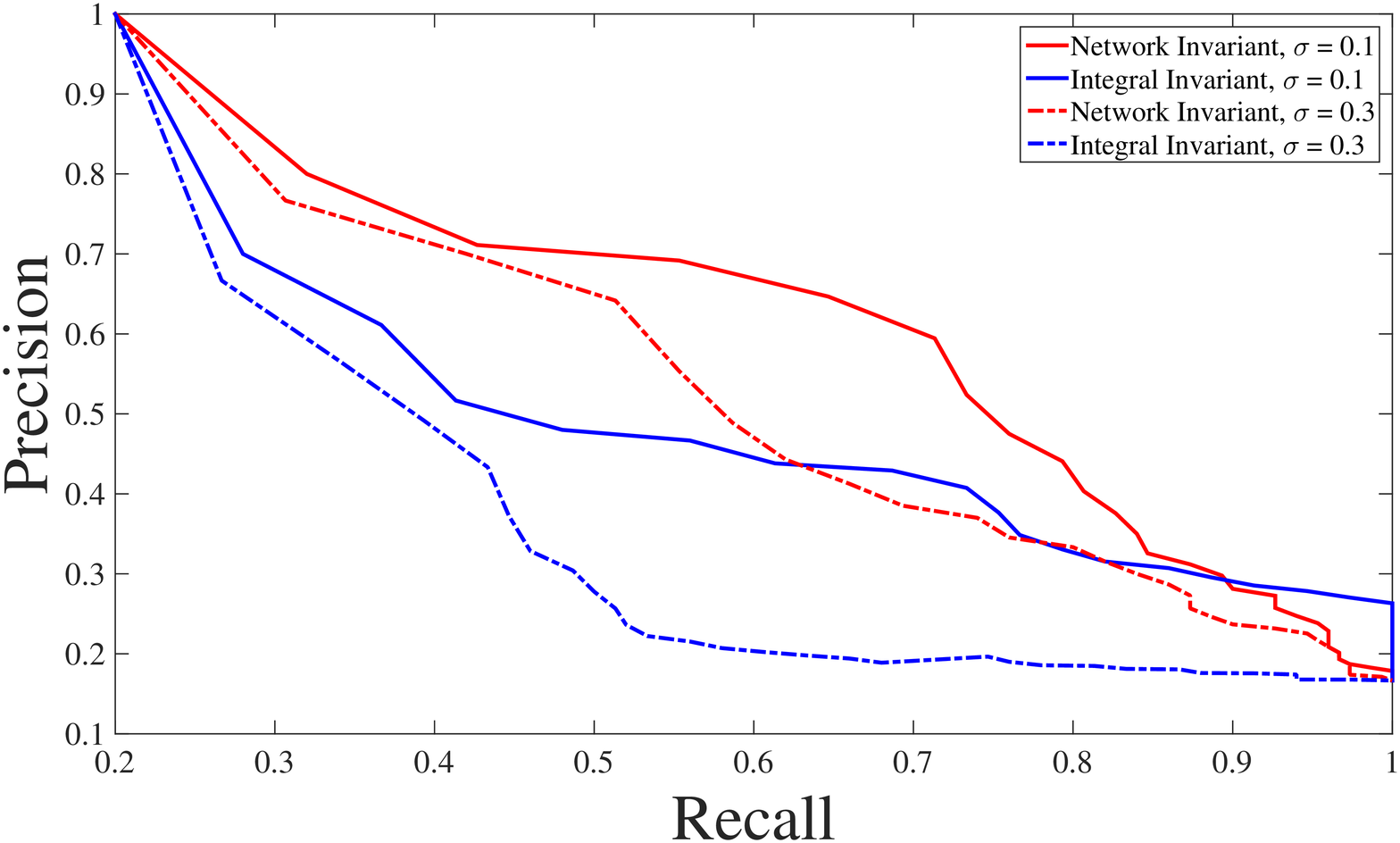}
\end{subfigure}
{\caption{$5$ shape contours of $6$ different categories and the shape retrieval 
 results for this set for different noise levels.}
\label{Retrieval_Experiment}}
\end{figure}

Ability to handle low signal to noise ratios and efficiency of computation are typical qualities
 desired in a geometric invariant. 
To test the numerical stability and robustness of the invariant signatures we designed two experiments. 
In the first experiment, we add increasing levels of zero-mean Gaussian noise to the curve and 
 compare the three  types of signatures: differential (Euclidean curvature),  
 integral (integral area invariant) and the output of our  network 
 (henceforth termed as network invariant) as shown in Figure \ref{Signatures_Noise}. 
Apart from adding noise, we also rotate the curve to obtain a better assessment of the Euclidean invariance property. 
In Figure \ref{Retrieval_Experiment}, we test descriptiveness of the signature under noisy conditions
 in a  shape retrieval task for a set of $30$ shapes with $6$ different categories. 
For every curve, we generate $5$ signatures at different scales for the integral and the network 
 invariant and use them as a representation for that shape. 
We use the Hausdorff distance as a distance measure (\cite{bronstein2008numerical}) between the two sets
 of signatures  to rank the shapes for retrieval. 
Figure \ref{Signatures_Noise} and \ref{Retrieval_Experiment} demonstrate the robustness of the 
 network especially at high noise levels. 

\begin{figure}[t]
\hspace{-0.8\textwidth}
\begin{subfigure}[h]{0.3\textwidth}
\centering
\includegraphics[scale=0.3]{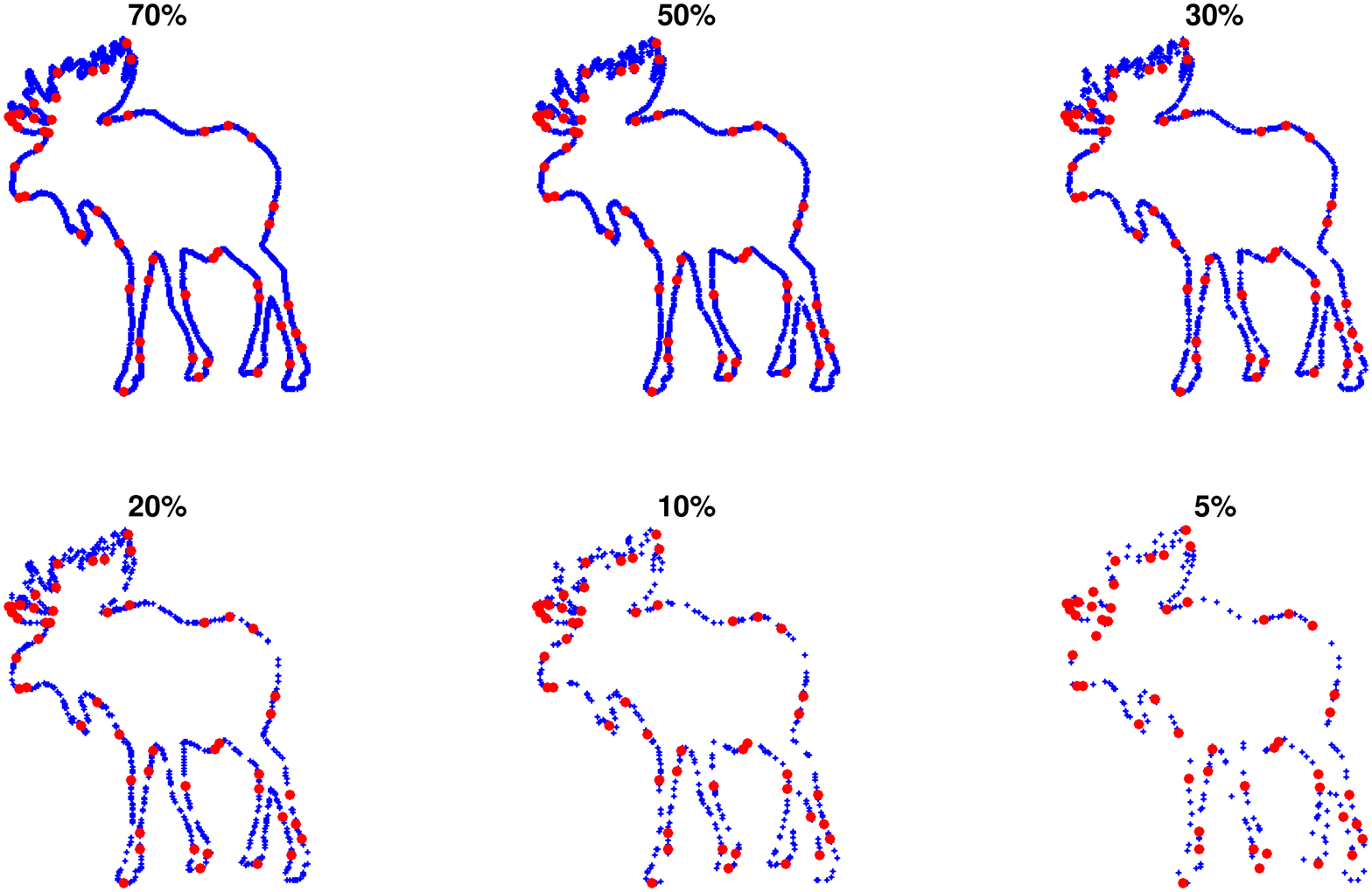}
\end{subfigure}

\hspace{-0.5\textwidth}
\begin{subfigure}[h]{0.3\textwidth}
\includegraphics[scale=0.23]{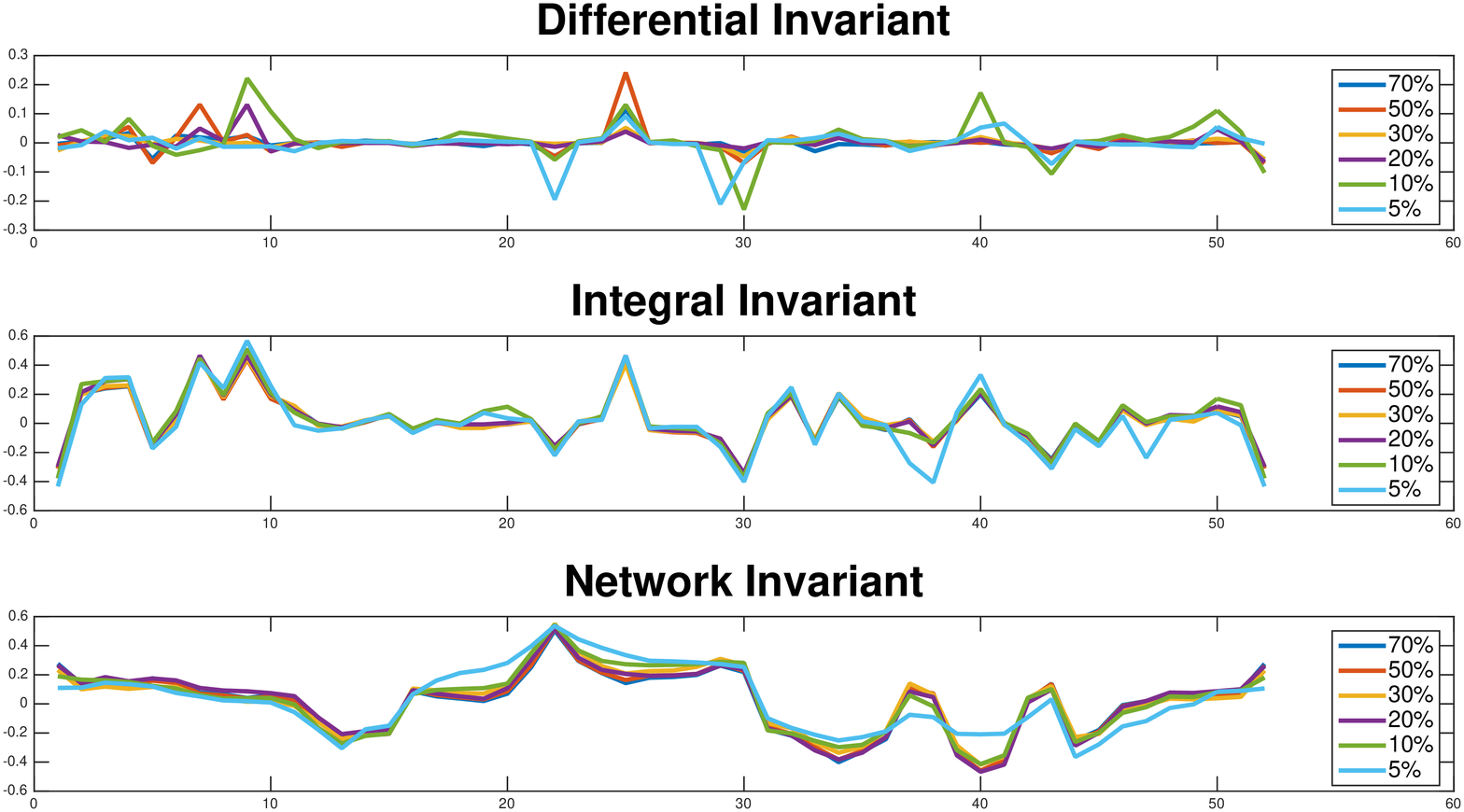}
\end{subfigure}
\caption{Testing robustness of signatures to different sampling conditions. 
        The signatures are evaluated at the \emph{fixed red} points on each contour 
         and the density and distribution of the blue points along the curve is varied 
          from $70\%$ to $5\%$ of the total number of points of a high resolution curve.} 
\label{Signatures_Sampling}
\end{figure}

In the second experiment, we decimate a high resolution contour at successive resolutions by randomly sub-sampling 
 and redistributing a set of its points (marked blue in Figure \ref{Signatures_Sampling}) and observe
  the signatures  at certain \emph{fixed} points (marked red in Figure \ref{Signatures_Sampling}) on
  the curve.  
Figure \ref{Signatures_Sampling} shows that the network is able to handle these changes in sampling and 
 compares  well with the integral invariant. 
Figures \ref{Signatures_Noise} and Figure \ref{Signatures_Sampling} represent behavior of
 geometric signatures  for two different tests: large noise for a moderate strength of signal and
 low signal for a moderate level of noise.
\section{Conclusion}\label{Conclusions}
We have demonstrated a method to learn geometric invariants of planar curves. 
Using just positive and negative examples of Euclidean transformations, we showed that
 a convolutional neural network is able to effectively discover and encode transform-invariant properties
 of curves while remaining numerically robust in the face of noise. 
By using a geometric context to the training process we were able to develop novel multi-scale 
 representations from a learning based approach without explicitly enforcing such behavior. 
As compared to a more axiomatic framework of modeling with differential geometry and engineering 
 with numerical analysis, we demonstrated a way of replacing this pipeline with a deep learning 
 framework which combines both these aspects. The non-specific nature of this framework can be seen as providing the groundwork for future deep learning data based problems in differential geometry.
\subsubsection*{Acknowledgments}
This project has received funding from the European Research Council (ERC) under the European Union’s
Horizon 2020 research and innovation program (grant agreement No 664800)
\bibliography{iclr2017_conference}
\bibliographystyle{iclr2017_conference}
\clearpage
\section{Appendix}
\begin{figure}[h]
\hspace{-1cm}
\begin{subfigure}{6cm}
\begin{tikzpicture} [thick, every node/.style={scale=0.85,font=\LARGE}]
\node(diff_0)[curvebox]{\includegraphics[scale = 0.1] {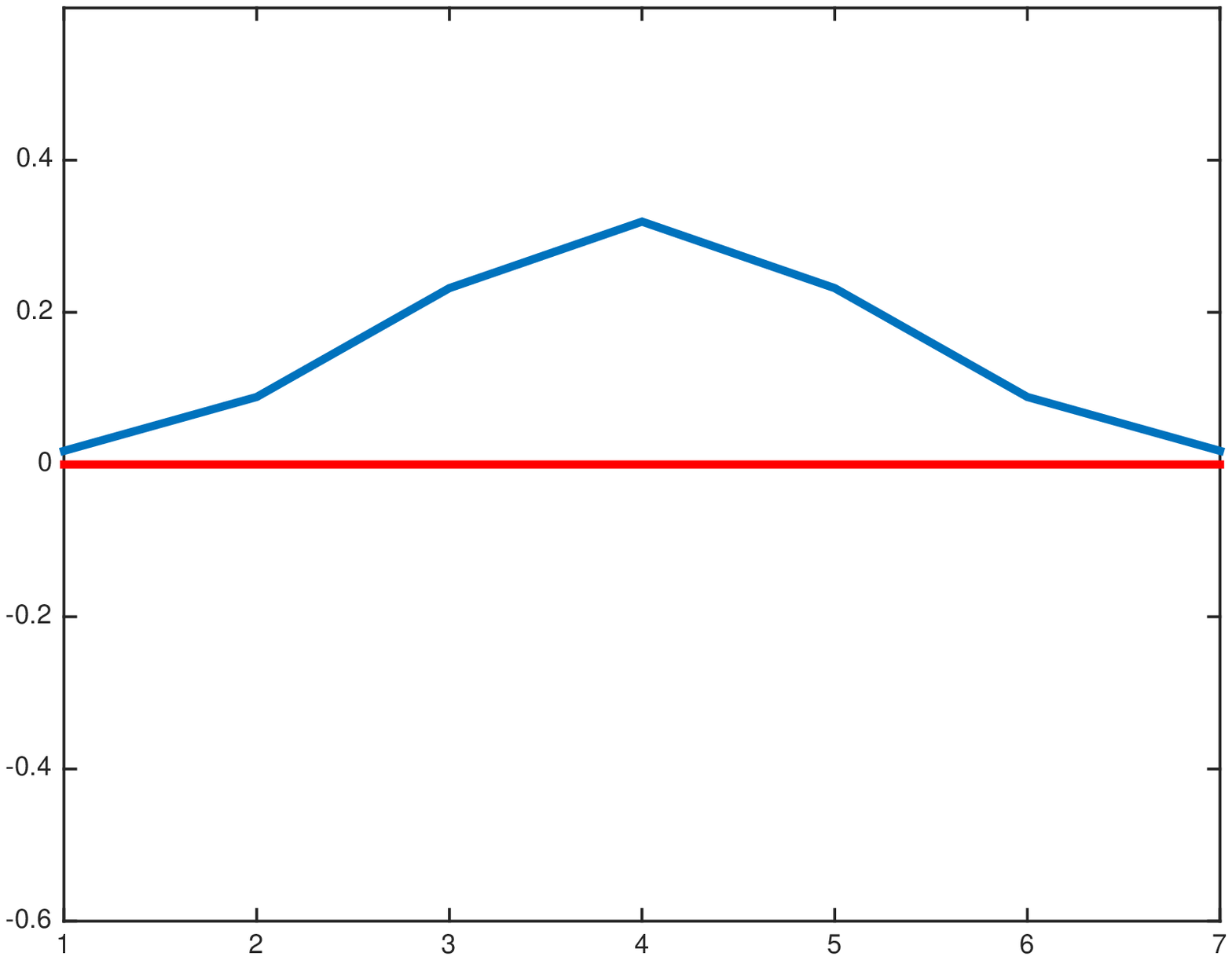}};

\node(diff_1)[curvebox, below of = diff_0, yshift = -0.75cm]{\includegraphics[scale = 0.1] {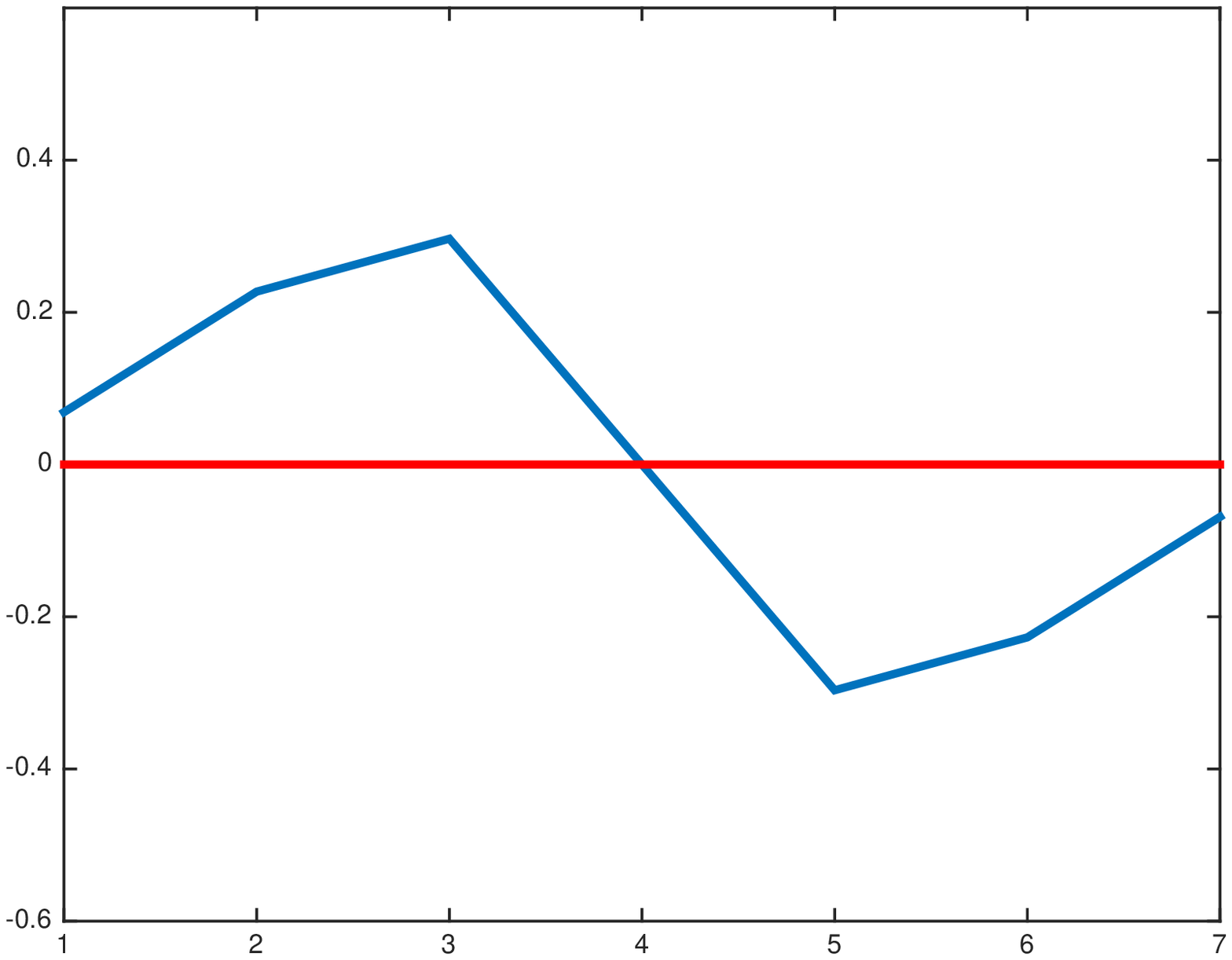}};

\node(diff_2)[curvebox, below of = diff_1, yshift = -0.75cm]{\includegraphics[scale = 0.1] {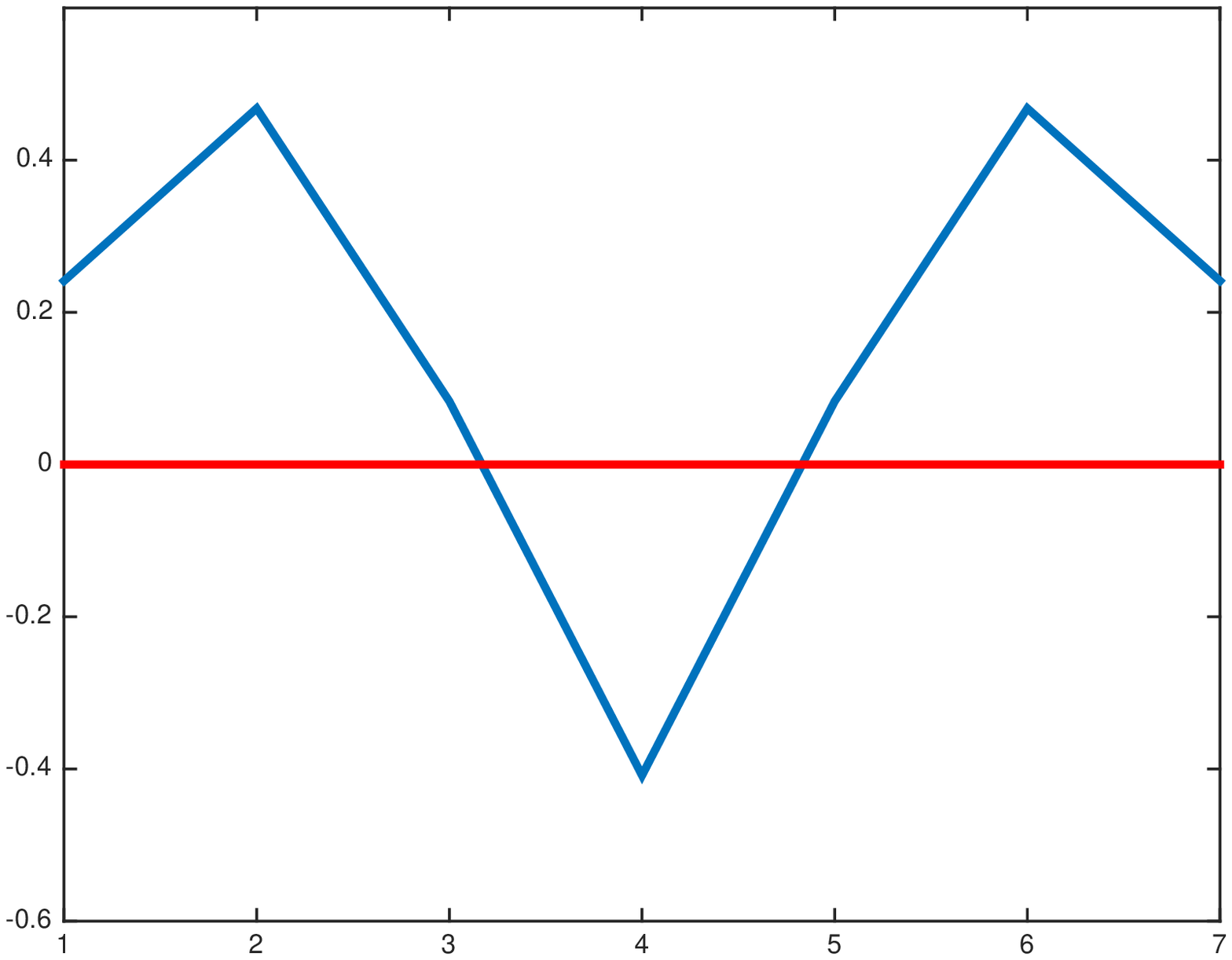}};

\node(text_0)[textbox, left of = diff_0, xshift = -2.5cm]{\large{$g(x,\sigma) = \frac{1}{\sigma \sqrt{2\pi}} \; e^{-\frac{x^2}{2 \sigma^2}}$}};

\node(text_1)[textbox, left of = diff_1, xshift = -2.5cm]{\large{$\frac{d}{dx} \; g(x,\sigma) = -\frac{x}{\sigma^3 \sqrt{2\pi}} \; e^{-\frac{x^2}{2 \sigma^2}}$}};

\node(text_2)[textbox, left of = diff_2, xshift = -2.5cm]{\large{$\frac{d^2}{dx^2} \; g(x,\sigma) = -\frac{\sigma^2 - x^2}{\sigma^5 \sqrt{2\pi}} \; e^{-\frac{x^2}{2 \sigma^2}}$}};
\end{tikzpicture}
\caption{}
\end{subfigure}
\begin{subfigure}{7.25cm}
\includegraphics[scale = 0.2] {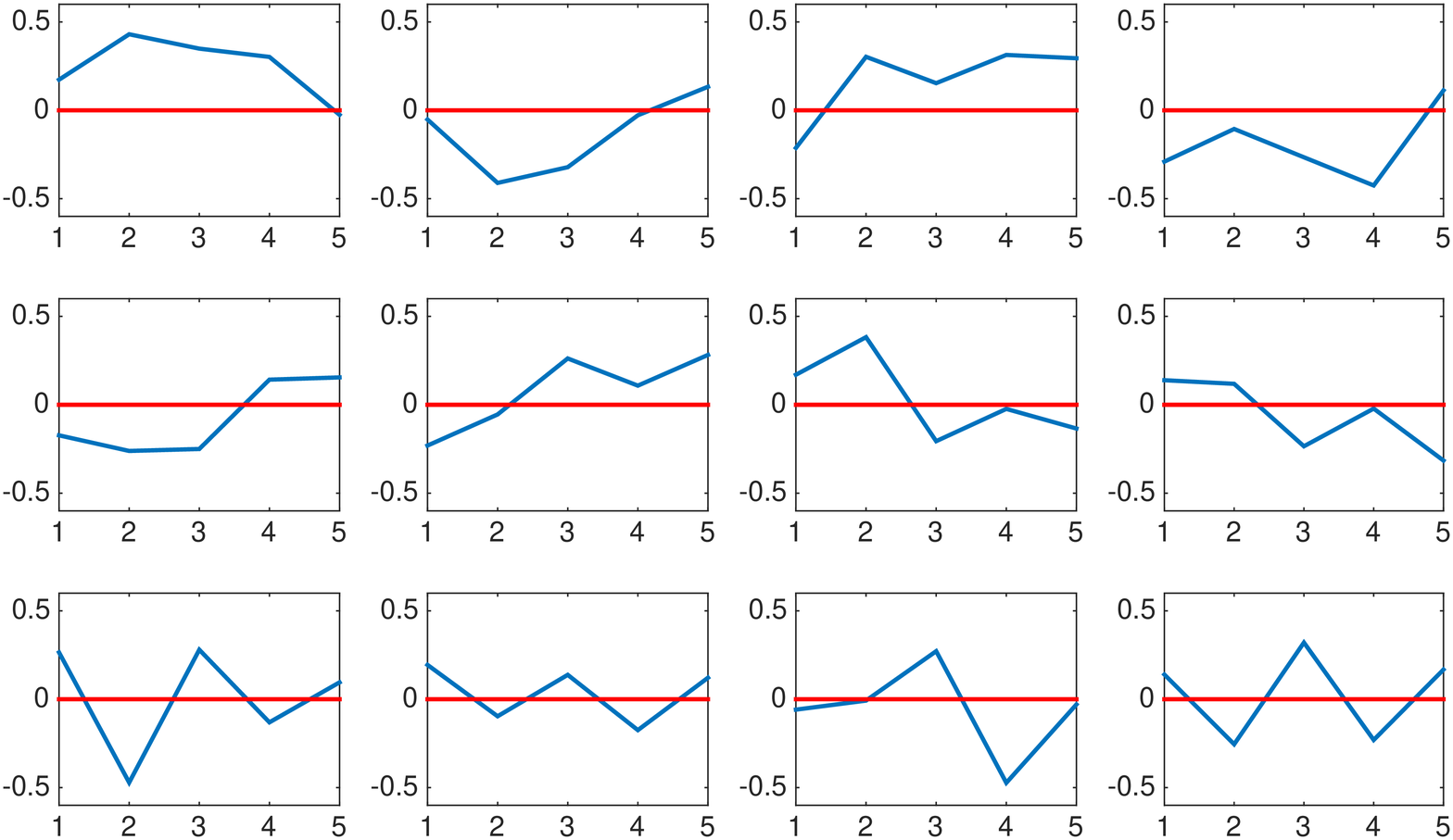}
\caption{}
\end{subfigure}
\caption{(a) Standard 1D Gaussian filters and its derivatives used for curvature and curvature scale space calculations. (b) Some of the filters from the first layer of the network proposed in this paper. One can interpret the shapes of the filters in (b) as derivative kernels which are learned from data and therefore adapted to its sampling conditions.}
\label{filters}
\end{figure}
\end{document}